\PassOptionsToPackage{unicode}{hyperref}
\PassOptionsToPackage{hyphens}{url}
\documentclass[
]{article}
\usepackage{xcolor}
\usepackage{amsmath,amssymb}
\usepackage{iftex}
\ifPDFTeX
  \usepackage[T1]{fontenc}
  \usepackage[utf8]{inputenc}
  \usepackage{textcomp} % provide euro and other symbols
\else % if luatex or xetex
  \usepackage{unicode-math} % this also loads fontspec
  \defaultfontfeatures{Scale=MatchLowercase}
  \defaultfontfeatures[\rmfamily]{Ligatures=TeX,Scale=1}
\fi
\usepackage{lmodern}
\ifPDFTeX\else
\fi
\IfFileExists{upquote.sty}{\usepackage{upquote}}{}
\IfFileExists{microtype.sty}{% use microtype if available
  \usepackage[]{microtype}
  \UseMicrotypeSet[protrusion]{basicmath} % disable protrusion for tt fonts
}{}
\makeatletter
\@ifundefined{KOMAClassName}{% if non-KOMA class
  \IfFileExists{parskip.sty}{%
    \usepackage{parskip}
  }{% else
    \setlength{\parindent}{0pt}
    \setlength{\parskip}{6pt plus 2pt minus 1pt}}
}{% if KOMA class
  \KOMAoptions{parskip=half}}
\makeatother
\usepackage{longtable,booktabs,array}
\usepackage{calc} % for calculating minipage widths
\usepackage{etoolbox}
\makeatletter
\patchcmd\longtable{\par}{\if@noskipsec\mbox{}\fi\par}{}{}
\makeatother
\IfFileExists{footnotehyper.sty}{\usepackage{footnotehyper}}{\usepackage{footnote}}
\makesavenoteenv{longtable}
\usepackage{graphicx}
\makeatletter
\newsavebox\pandoc@box
\newcommand*\pandocbounded[1]{% scales image to fit in text height/width
  \sbox\pandoc@box{#1}%
  \Gscale@div\@tempa{0.68\textheight}{\dimexpr\ht\pandoc@box+\dp\pandoc@box\relax}%
  \Gscale@div\@tempb{\linewidth}{\wd\pandoc@box}%
  \ifdim\@tempb\p@<\@tempa\p@\let\@tempa\@tempb\fi% select the smaller of both
  \ifdim\@tempa\p@<\p@\scalebox{\@tempa}{\usebox\pandoc@box}%
  \else\usebox{\pandoc@box}%
  \fi%
}
\def\fps@figure{htbp}
\makeatother
\providecommand{\tightlist}{%
  \setlength{\itemsep}{0pt}\setlength{\parskip}{0pt}}
\usepackage[section]{placeins}
\usepackage{float}
\usepackage{caption}
\usepackage{bookmark}
\IfFileExists{xurl.sty}{\usepackage{xurl}}{} % add URL line breaks if available
\hypersetup{
  hidelinks,
  pdfcreator={LaTeX via pandoc}}

\author{}
\date{}

\begin{document}

\section{RegNetAgents: A Multi-Agent Framework for Cross-Network
Regulatory Driver Identification in Cancer
Genomics}\label{regnetagents-a-multi-agent-framework-for-cross-network-regulatory-driver-identification-in-cancer-genomics}

\textbf{Author:} Jose A. Bird (ORCID: 0009-0006-2744-0606)
\textbf{Affiliation:} Independent Researcher \textbf{Correspondence:}
jbird@birdaisolutions.com

\begin{center}\rule{0.5\linewidth}{0.5pt}\end{center}

\subsection{Abstract}\label{abstract}

We introduce RegNetAgents, an AI-oriented multi-agent framework for
structured, query-driven regulatory candidate identification across
heterogeneous gene regulatory networks. The system enables unified
analysis of bulk tumor and single-cell-derived ARACNe networks by
integrating TCGA-derived cancer networks with large-scale single-cell
regulatory networks from the GREmLN project. RegNetAgents enables
source-aware reasoning across multiple network contexts, labeling each
candidate regulator by the network or networks in which it appears.

For a given focal gene, the framework finds its regulators in both
networks and labels each by source (Both, TCGA-only, GREmLN-only), flags
those that are known cancer driver genes (IntOGen), and, for
tumor-network regulators, gives the mode of action (MoA; activating or
repressive). The system is implemented as a multi-agent LangGraph
state-graph workflow, accessible through a unified Python API and Model
Context Protocol (MCP) client, operating as a downstream analytical
layer over precomputed regulatory networks rather than a network
inference method. For example, a single query for CTNNB1 in BRCA returns
two tumor-specific (TCGA-only) driver-gene regulators, DDR2 and IL6ST,
both with activating MoA. A second module, rule-based by default,
profiles shortlisted candidates by network topology; it is not evaluated
here.

Across twelve breast cancer (BRCA) and thirteen colorectal cancer (COAD)
driver genes, we compared each gene's tumor-network-only (TCGA-only)
regulators with the TCGA-only regulators of random non-driver genes.
Regulators as a group are about 2.7-fold richer in driver genes than
genes overall, so almost any gene's regulators look enriched when tested
against all genes. We therefore used random genes as the baseline. In
COAD, the cancer genes' regulators include modestly but consistently
more IntOGen driver genes than random genes' regulators do (nominal p =
0.012--0.036 across five random samples); in BRCA the difference is
borderline (p = 0.048--0.083). Housekeeping and non-driver control genes
show no such excess in the tumor-network tier, and the same comparison
for GREmLN-only regulators shows none (p \ensuremath{\geq} 0.20).

Code is available at: \url{https://github.com/jab57/RegNetAgents}

\textbf{Keywords:} gene regulatory network, cancer genomics, ARACNe,
single-cell transcriptomics, multi-agent systems, LangGraph, regulatory
candidate identification, mode-of-action, TCGA, IntOGen

\textbf{Revision note (v3).} This version corrects and replaces the
analysis of versions 1--2. (i) The cancer-gene reference set changes
from OncoKB to the IntOGen compendium {[}Martínez-Jiménez et al.,
2020{]} (release 2024.09.20, CC0), so that the reference can be
distributed with the code, making every result reproducible, and so that
it matches the driver labels RegNetAgents reports; all enrichment
results are recomputed. (ii) A comparison with random genes is added; it
replaces the permutation control of v1--v2, which reproduced the Fisher
p-value (same null hypothesis) and so was not an independent check. It
shows that the GREmLN-only enrichment reported in v1--v2 is not specific
to cancer genes, while TCGA-only candidates retain a modest excess of
driver genes over random genes (consistent in COAD, borderline in BRCA),
smaller than the all-gene comparison suggested. The combined p-values of
v1--v2 (p \textless{} 0.0001) are therefore no longer reported; Stouffer
Z is given as a descriptive summary only. Per-gene results also depend
on the reference set, so differences from v1--v2 reflect both (i) and
(ii). (iii) The focal gene panels, selected for v1 from OncoKB-annotated
driver genes, are kept; CDH1 and RNF43, which v1--v2 excluded because of
a gene-identifier bug (fixed in v1.3.0), are restored, and PABPC1 is
dropped from the neutral controls (not replaced) because it is an
IntOGen driver. BRCA1, described in v1--v2 as absent from the TCGA BRCA
network, is present there but has no regulators, so it remains
untestable in the TCGA-only tier. (iv) The domain-agent table of v1--v2
(Table 10) is removed because its ratings were produced in the optional
LLM mode rather than by the deterministic rules described. (v) Several
statements and references are corrected, including softened claims about
MoA literature support, PTEN and cross-network coverage, the
\texttt{aracne.networks} author attribution, and other reference
details. For reference, versions 1--2 used the OncoKB cancer gene list
(source: OncoKB, Memorial Sloan Kettering Cancer Center {[}Chakravarty
et al., 2017; Suehnholz et al., 2024{]}; see its Terms of Use,
\url{https://www.oncokb.org/terms}).

\begin{center}\rule{0.5\linewidth}{0.5pt}\end{center}

\subsection{1. Introduction}\label{introduction}

Identifying candidate regulatory drivers of cancer genes is a central
problem in cancer systems biology {[}Califano \& Alvarez, 2017{]}.
Transcription factors that regulate key oncogenes or tumor suppressor
genes --- and whose activity is supported by bulk-tumor network
inference --- are candidates for follow-up study. Distinguishing
regulators supported by tumor networks from those broadly active across
cell types requires querying multiple network data sources, classifying
regulators by source, filtering against cancer driver resources, and
annotating mode-of-action directionality. To our knowledge, no tool
currently integrates these steps.

RegNetAgents integrates two complementary ARACNe-based network
resources. Single-cell ARACNe networks from the GREmLN project {[}Zhang
et al., 2025{]} are inferred with ARACNe-AP {[}Lachmann et al., 2016{]}
from CELLxGENE Census data {[}CZI Cell Science Program et al., 2025{]}
(the GREmLN corpus spans 11M cells across 162 cell types), and represent
population-averaged regulatory relationships within each cell type
across a large multi-study atlas; these networks should not be
interpreted as strictly normal tissue baselines. Bulk-tumor ARACNe
networks from the Bioconductor \texttt{aracne.networks} package
{[}Giorgi, 2025; Giorgi \& Alvarez, 2026{]} are derived from TCGA
RNA-seq data and capture the aggregate regulatory topology of each
cancer type, with per-edge mode-of-action (MoA) annotations encoding
activating or repressive relationships. Together these sources offer
distinct perspectives on gene regulation --- but querying both,
classifying regulators by source, filtering against cancer driver
resources, and annotating MoA directionality currently requires multiple
independent steps and manual curation. Cross-network differences reflect
both technical factors (single-cell vs.~bulk inference, differing gene
coverage and mutual-information (MI) thresholds) and biological factors,
and should not be interpreted as a clean normal-vs-tumor contrast.

Existing GRN analysis tools address parts of this problem but leave a
gap. ARACNe {[}Basso et al., 2005{]} and its successors are designed for
network construction, not for querying pre-built networks or classifying
regulators by source. VIPER {[}Alvarez et al., 2016{]} infers
transcription factor activity from expression signatures within a single
network and does not compare regulator sets across independently derived
networks. PANDA {[}Glass et al., 2013{]} models network differences but
requires user-supplied expression matrices. SCENIC {[}Aibar et al.,
2017{]} infers regulons de novo from single-cell expression data and
does not operate on pre-built networks or classify regulators by network
source. None provides dual-source classification across two
independently inferred ARACNe topologies, automatic cancer-driver
filtering, or MoA directionality annotation in a single query. The
novelty of RegNetAgents lies in this integration layer --- combining two
independently built network databases, a cancer driver reference, and
MoA annotation into a single reproducible query --- rather than in the
underlying network inference or set-comparison operations, which are
straightforward applications of existing methods.

RegNetAgents addresses this gap primarily through
\texttt{compare\_network\_contexts}, which produces source-labeled,
MoA-annotated, driver-filtered regulatory candidate lists --- distilling
a focal gene's regulatory neighborhood into an ordered, annotated
shortlist in a single query. A second operation,
\texttt{comprehensive\_gene\_analysis}, provides rule-based,
topology-derived profiling of selected candidates; it is described
briefly but not evaluated in this paper. Both are accessible through a
Python API and a Model Context Protocol (MCP) client. Source labels are
derived by querying GREmLN project ARACNe and TCGA bulk-tumor ARACNe
networks and classifying each regulator in the union of both sets by
network of origin; MoA directionality is drawn from per-edge TCGA
network annotations. The empirical question of this paper is whether the
candidates produced for cancer driver genes are enriched for known
driver genes beyond what the candidates of random genes show. Across
twelve BRCA and thirteen COAD focal genes, TCGA-only candidates show a
modest excess of driver genes over random genes in COAD and a borderline
one in BRCA, whereas GREmLN-only candidates show none.

\begin{center}\rule{0.5\linewidth}{0.5pt}\end{center}

\subsection{2. Methods}\label{methods}

\subsubsection{2.1 RegNetAgents
Architecture}\label{regnetagents-architecture}

RegNetAgents is a Python package for automated downstream analysis of
pre-computed gene regulatory networks. It is installable via pip and
available at \url{https://github.com/jab57/RegNetAgents} (v1.3.0, archived on
Zenodo, DOI: 10.5281/zenodo.23215310). Users interact through either a
Python API or a Model Context Protocol (MCP) client, enabling
natural-language queries without writing code.

The system exposes two analytical operations.
\texttt{compare\_network\_contexts} is a deterministic, rule-based
operation that queries both network backends for a focal gene and
classifies regulators by source; all results in this paper come from it.
\texttt{comprehensive\_gene\_analysis} is a LangGraph state-graph
workflow {[}LangChain AI, 2024{]} coordinating four domain agents. In
the default rule-based mode, each agent produces a deterministic
assessment (high/moderate/low) from network topology metrics against
empirically derived per-network percentile thresholds (75th and 90th;
rules in the table below). A gene is classified as a hub regulator when
its out-degree (downstream target count) exceeds the 90th percentile
threshold for the queried network (GREmLN cell-type or TCGA
cancer-type), computed across all nodes in that network. Out-degree is a
gene's number of targets and in-degree its number of upstream
regulators; percentiles refer to the queried network.

\noindent\begin{minipage}{\linewidth}
{\def\LTcaptype{none}
\begin{tabular}{@{}
  >{\raggedright\arraybackslash}p{(\linewidth - 6\tabcolsep) * \real{0.2500}}
  >{\raggedright\arraybackslash}p{(\linewidth - 6\tabcolsep) * \real{0.2500}}
  >{\raggedright\arraybackslash}p{(\linewidth - 6\tabcolsep) * \real{0.2500}}
  >{\raggedright\arraybackslash}p{(\linewidth - 6\tabcolsep) * \real{0.2500}}@{}}
\toprule\noalign{}
\begin{minipage}[b]{\linewidth}\raggedright
Domain agent
\end{minipage} & \begin{minipage}[b]{\linewidth}\raggedright
High
\end{minipage} & \begin{minipage}[b]{\linewidth}\raggedright
Moderate
\end{minipage} & \begin{minipage}[b]{\linewidth}\raggedright
Low
\end{minipage} \\
\midrule\noalign{}
Oncogenic potential & Hub (out-degree \textgreater{} 90th percentile) &
Out-degree \textgreater{} 75th percentile & Otherwise \\
Druggability & Hub, or out-degree \textgreater{} 75th percentile with
in-degree \ensuremath{\leq} 75th & Out-degree \textgreater{} 75th with in-degree
\textgreater{} 75th; or other genes with targets and in-degree \ensuremath{\leq} 90th &
No targets; or in-degree \textgreater{} 90th with out-degree \ensuremath{\leq} 75th \\
Clinical actionability & Hub, or in-degree \textgreater{} 75th
percentile & Otherwise & --- \\
Network vulnerability & \emph{Critical:} hub & \emph{Important:}
out-degree \textgreater{} 75th percentile & \emph{Minimal:} otherwise \\
\bottomrule\noalign{}
\end{tabular}
}
\end{minipage}

All four assessments are computed from network connectivity only; the
domain names describe the intended use of each signal.

In an optional LLM mode, a language model assigns the domain assessments
instead of the rules; no result in this paper uses LLM mode.
\texttt{comprehensive\_gene\_analysis} accepts either a GREmLN cell-type
or a TCGA cancer-type network as topology source.

The framework integrates two network backends:

\begin{itemize}
\tightlist
\item
  \textbf{GREmLN project ARACNe networks} {[}Zhang et al., 2025{]}:
  population-averaged regulatory networks for 10 cell types (1
  epithelial and 9 immune/blood cell types), inferred with ARACNe-AP
  {[}Lachmann et al., 2016{]} from CELLxGENE Census data {[}CZI Cell
  Science Program et al., 2025{]} (the GREmLN corpus spans 11M cells
  across 162 cell types). These are the pre-built ARACNe-AP network
  files distributed by the GREmLN project via the GREmLN Quickstart
  Tutorial (CZI Virtual Cells Platform) --- distinct from the GREmLN
  foundation model itself (model code: github.com/czi-ai/GREmLN).
  RegNetAgents converts each network file into a PKL lookup cache
  (\path{scripts/build_network_cache.py}) and translates Ensembl
  gene IDs to symbols at query time using a cache it builds from
  MyGene.info {[}Xin et al., 2016{]}. The GREmLN epithelial\_cell
  network encompasses 14,621 symbol-resolved genes (of 14,628 unique
  Ensembl IDs) and 1,926 regulators.
\item
  \textbf{TCGA ARACNe} {[}Giorgi, 2025{]}: tumor-state networks for 14
  epithelial-origin cancer types from Bioconductor
  \texttt{aracne.networks} v1.36.0, installed from the package authors'
  Zenodo record {[}Giorgi \& Alvarez, 2026{]} by a setup script. Each
  edge carries a mode-of-action (MoA) value between \ensuremath{-}1 and +1, derived
  from regulator--target expression correlation (positive = activating,
  negative = repressive). The GREmLN networks provide only
  mutual-information edge weights, so MoA is available for TCGA networks
  only.
\end{itemize}

The \texttt{compare\_network\_contexts} operation queries a focal gene
in both a specified GREmLN cell type and a specified TCGA cancer type,
then classifies each regulator as: \textbf{Both} (present in both
networks), \textbf{GREmLN-only} (single-cell ARACNe-specific), or
\textbf{TCGA-only} (bulk-tumor ARACNe-specific). Overlap between the two
regulator sets is measured by the Jaccard similarity J = \textbar GREmLN
\(\cap\) TCGA\textbar{} / \textbar GREmLN \(\cup\) TCGA\textbar,
reported as \texttt{conserved\_\allowbreak{}fraction} (Table 2); 1 \ensuremath{-} J is the
context-specificity, near 1 when the two networks share almost no
regulators. Regulators in the intersection correspond to the
\textbf{Both} source category.

\subsubsection{2.2 Reference Gene Set}\label{reference-gene-set}

\textbf{IntOGen} {[}Martínez-Jiménez et al., 2020{]}: the IntOGen
Compendium of Mutational Cancer Driver Genes, release 2024.09.20,
distributed under CC0 1.0 (public domain) and bundled with RegNetAgents
(\path{regnetagents/reference_data/intogen_drivers.tsv}). The
reference set is every gene in the compendium --- 633 genes, pan-cancer,
of any role (oncogene, tumor suppressor, mixed, or ambiguous); no role
or tissue subsets were used. After intersecting with each network's gene
universe, 620 IntOGen genes were available as reference for TCGA BRCA,
622 for TCGA COAD, and 575 for the GREmLN epithelial\_cell network.
IntOGen identifies driver genes from somatic mutation data, including
TCGA cohorts, whereas the TCGA networks are inferred from expression
data; the two therefore use different data types, but they partly derive
from the same patients (see Limitations).

The reference set and the analysis design (panels, controls, and
statistics) were fixed before any IntOGen-based analysis was run.

\subsubsection{2.3 Focal Gene Panels}\label{focal-gene-panels}

The focal gene panels were selected from OncoKB-annotated cancer driver
genes {[}Chakravarty et al., 2017; Suehnholz et al., 2024{]} and fixed
before the analysis. For BRCA, thirteen focal genes span major oncogenic
programs --- cell cycle regulation (CCND1, RB1), PI3K-AKT signaling
(PIK3CA, PTEN), hormone receptor signaling (ESR1, GATA3), Wnt/\ensuremath{\beta}-catenin
signaling (CTNNB1), DNA damage response (BRCA1, BRCA2), cell adhesion
(CDH1), and established breast cancer drivers (MYC, ERBB2, TP53). BRCA1
has no regulators in the TCGA BRCA network (it appears there only as a
regulator of other genes), so it is not testable in the TCGA-only tier,
leaving twelve: \textbf{TP53}, \textbf{MYC}, \textbf{CTNNB1},
\textbf{CCND1}, \textbf{BRCA2}, \textbf{PIK3CA}, \textbf{PTEN},
\textbf{RB1}, \textbf{ERBB2}, \textbf{ESR1}, \textbf{GATA3}, and
\textbf{CDH1}. BRCA1 is included in the GREmLN-only analysis. PIK3CA has
fewer than three GREmLN-only candidates in either cancer type and is
therefore not tested in the GREmLN-only tier (Section 2.4).

For COAD, thirteen focal genes were selected based on mutation frequency
and pathway representation in colorectal cancer {[}Cancer Genome Atlas
Network, 2012{]}: \textbf{TP53, MYC, CTNNB1, CCND1}, \textbf{KRAS},
\textbf{APC}, \textbf{SMAD4}, \textbf{BRAF}, \textbf{PIK3CA},
\textbf{PTEN}, \textbf{FBXW7}, \textbf{TCF7L2}, and \textbf{RNF43}. Six
of these (TP53, MYC, CTNNB1, CCND1, PIK3CA, and PTEN) are also in the
BRCA panel.

All focal genes are IntOGen drivers.

The reference cell type for all comparisons was
\texttt{epithelial\_cell}, the only epithelial network among the ten
GREmLN networks and the only one with broad gene coverage (14,628 genes;
the nine immune and blood cell networks cover 411--3,086), and therefore
the closest available match to epithelial-origin carcinomas such as BRCA
and COAD. It is not tissue-matched: this network is a pan-tissue
population average derived from the CELLxGENE Census --- a large-scale
atlas aggregating diverse studies --- and should be interpreted as a
broad single-cell reference context rather than a strictly normal tissue
baseline.

\subsubsection{2.4 Candidate List
Construction}\label{candidate-list-construction}

For each focal gene, \texttt{compare\_network\_contexts} takes the union
of all regulators found in either network, classifies each by network
source (TCGA-only, GREmLN-only, or Both), and returns MoA direction for
TCGA-sourced entries (activating if MoA \textgreater{} 0, repressive if
MoA \textless{} 0). MoA is derived from the expression correlation
between regulator and target (positive = activating, negative =
repressive); it indicates the direction of co-expression, not a
demonstrated mechanism or the direction of regulation. Focal genes with
fewer than three regulators in a given tier (counted before IntOGen
filtering) were excluded from Fisher's exact test for that tier, as 2\ensuremath{\times}2
tables with very sparse cells produce unreliable odds ratio estimates.
The analysis pipeline
(\path{scripts/experiment_regulatory_candidates.py}) then filters
regulators from all source tiers against IntOGen to produce a
source-labeled candidate list. Table headers use ``tumor-selective'' as
a synonym for TCGA-only; it is an operational label based on network
source, not a claim of tumor-specific biological function. Each entry
records: (i) network source; (ii) IntOGen role (oncogene, tumor
suppressor, mixed, or ambiguous); and (iii) MoA direction for
TCGA-sourced entries. Candidates are ordered with Both-source entries
first, followed by TCGA-only (by MoA, activating before repressive),
then GREmLN-only. This ordered, source-labeled list is the primary
deliverable (Figure 1).

\subsubsection{2.5 Statistical Framework}\label{statistical-framework}

\textbf{Per-gene enrichment.} Enrichment of each focal gene's candidates
for IntOGen genes was assessed with Fisher's exact test (one-tailed,
alternative = ``greater''). For TCGA-only candidates, all genes in the
queried TCGA cancer-type network served as background; for GREmLN-only
candidates, the GREmLN epithelial\_cell gene universe (14,621 genes) was
used. Because the GREmLN network is pan-tissue rather than
cancer-type-specific, the same background is used for both BRCA and COAD
focal genes. Benjamini-Hochberg FDR correction was applied across focal
genes within each cancer type and tier. Stouffer's weighted Z-score
combined the one-tailed p-values across focal genes, with weights
proportional to candidate set size; a gene with no IntOGen gene among
its candidates (p = 1) contributes Z = 0. Z is reported as a descriptive
panel summary; its significance is assessed only against random genes
(below).

\textbf{Comparison with random genes.} Regulators are themselves
enriched for driver genes: 8.5\% of TCGA BRCA and COAD regulators are
IntOGen genes, against 3.1--3.2\% of all genes (2.7-fold), and 10.4\% of
GREmLN regulators against 3.9\% of the GREmLN universe (2.7-fold). Any
gene's regulator set therefore tends to look ``enriched'' against an
all-gene background. To test whether the focal genes' candidates exceed
this baseline, each tier test was run for 300 random genes per sample
(present in both networks; not IntOGen genes; not focal or control
genes), for five independent random samples (seeds 42, 1, 2, 3, 4).
Known driver genes were excluded from the random pool so that the
baseline represents genes with no known cancer-driver role, the contrast
the comparison is meant to test. As a result, the comparison reflects
driver genes in general rather than the chosen focal genes specifically
(see Limitations). Random genes with fewer than three regulators in a
tier were skipped, as for focal genes, leaving 296--300 testable per
sample in the TCGA-only tier and 210--239 in the GREmLN-only tier. The
focal panel was compared with the random genes by the mean
Haldane-corrected log odds ratio (0.5 added to each cell, so the measure
stays finite when a count is zero), using a one-sided two-sample
permutation test (5,000 permutations per sample). In each permutation,
the focal and random genes' log odds ratios were pooled and randomly
reassigned to the two groups, and the difference in group means was
recomputed; the p-value is the proportion of the 5,000 permuted
differences at least as large as the observed one, counting the observed
value itself (so the smallest possible p is 1/5,001). We report the
range of p-values across the five samples. For readability, effect sizes
are given as the geometric-mean OR (the exponential of the mean log odds
ratio). As a robustness check, the same comparison was repeated with a
regulator-only background (all regulators in the respective network),
against which random genes show much weaker enrichment (geometric-mean
OR 1.1--1.2 for TCGA-only, 1.7--2.0 for GREmLN-only). Separately from
the permutation test, we also report the median weighted Stouffer Z of
2,000 random panels per sample (each panel is a draw, without
replacement, of as many testable random genes as there are testable
focal genes, combined in the same way as the focal panel), only to show
what Z values arise by chance; no p-value from this comparison is
reported.

\subsubsection{2.6 Validation Controls and
Reproducibility}\label{validation-controls-and-reproducibility}

\textbf{Network scale.} The two network backends differ in scale: the
GREmLN epithelial\_cell network contains 1,926 regulators versus 6,052
in TCGA BRCA (\textasciitilde3\ensuremath{\times} difference), which biases classification
toward TCGA-only. Regulators absent from the GREmLN network, or present
but never acting as regulators there, are necessarily classified as
TCGA-only, reflecting network coverage rather than biological
exclusivity.

\textbf{Negative controls.} Two control panels were analyzed identically
to the focal panels. Housekeeping genes (ACTB, GAPDH, HPRT1, LDHA, TUBB)
test whether candidate sets of this size produce enrichment without
cancer-driver biology; they produce TCGA-only candidate sets of
comparable size to the focal genes (10--42 genes). Tumor-expressed
non-driver genes (``neutral controls''; FASN, PCNA, PKM, VIM) test
whether tumor-network membership alone produces enrichment; they produce
substantial TCGA-only candidate sets (8--39 genes).

\textbf{Determinism and reproducibility.} All enrichment statistics and
candidate selection are deterministic and independent of any language
model. All analyses were performed using RegNetAgents v1.3.0 (Zenodo
DOI: 10.5281/zenodo.23215310). The complete analysis script,
\path{scripts/experiment_regulatory_candidates.py}, is included in
the public repository and reproduces all tables, figures, and
supplementary tables in this paper. All other inputs are bundled; the
TCGA networks are installed with
\path{scripts/setup_tcga_networks.py}, which checks each downloaded
file and each rebuilt network against SHA-256 checksums recorded from
the files used in this paper, so every install matches the networks
analyzed here. All per-gene and combined statistics were independently
recomputed from the raw network files by separately written code, with
identical results; the random-gene comparison was independently repeated
with a different sampling scheme (all eligible random genes), with
consistent conclusions.

\begin{center}\rule{0.5\linewidth}{0.5pt}\end{center}

\subsection{3. Results}\label{results}

\subsubsection{3.1 Source-Labeled Candidate Lists: Output
Characterization and Cross-Network
Overlap}\label{source-labeled-candidate-lists-output-characterization-and-cross-network-overlap}

For each focal gene, \texttt{compare\_network\_contexts} produces a
source-labeled list of IntOGen-overlapping regulators classified by
network source: TCGA-only (bulk-tumor ARACNe, with MoA directionality),
GREmLN-only (single-cell ARACNe), or Both. Figure 2 shows the candidate
counts by network source for BRCA and COAD focal genes.

The CTNNB1 BRCA panel illustrates the output of a single query (Table
1). Of the 10 TCGA-only regulators, 2 are IntOGen driver genes, and both
carry activating MoA annotations (DDR2 and IL6ST, MoA \ensuremath{\approx} +1).

\noindent\begin{minipage}{\linewidth}
\textbf{Table 1. CTNNB1 BRCA candidate shortlist from a single
\texttt{compare\_network\_contexts} query (TCGA-only IntOGen-overlapping
regulators).}
\par\medskip
{\def\LTcaptype{none}
\begin{tabular}{@{}lllll@{}}
\toprule\noalign{}
Regulator & Source & IntOGen role & MoA & Direction \\
\midrule\noalign{}
DDR2 & TCGA-only & Tumor suppressor & \ensuremath{\approx} +1.00 & Activating \\
IL6ST & TCGA-only & Oncogene & \ensuremath{\approx} +1.00 & Activating \\
\bottomrule\noalign{}
\end{tabular}
}
\end{minipage}

Among all TCGA-only IntOGen-overlapping candidates, 28 of 39 in BRCA
(72\%) and 34 of 40 in COAD (85\%) carried activating MoA annotations
(MoA \textgreater{} 0; Table 2), with the remainder repressive.

\noindent\begin{minipage}{\linewidth}
\textbf{Table 2. Cross-network overlap and MoA distribution summary (all
focal gene--cancer type pairs).}
\par\medskip
{\def\LTcaptype{none}
\begin{tabular}{@{}
  >{\raggedright\arraybackslash}p{(\linewidth - 10\tabcolsep) * \real{0.11}}
  >{\raggedright\arraybackslash}p{(\linewidth - 10\tabcolsep) * \real{0.17}}
  >{\raggedright\arraybackslash}p{(\linewidth - 10\tabcolsep) * \real{0.15}}
  >{\raggedright\arraybackslash}p{(\linewidth - 10\tabcolsep) * \real{0.27}}
  >{\raggedright\arraybackslash}p{(\linewidth - 10\tabcolsep) * \real{0.14}}
  >{\raggedright\arraybackslash}p{(\linewidth - 10\tabcolsep) * \real{0.16}}@{}}
\toprule\noalign{}
\begin{minipage}[b]{\linewidth}\raggedright
Cancer type
\end{minipage} & \begin{minipage}[b]{\linewidth}\raggedright
Analyzable focal genes
\end{minipage} & \begin{minipage}[b]{\linewidth}\raggedright
Pairs with conserved\_\allowbreak{}fraction = 0
\end{minipage} & \begin{minipage}[b]{\linewidth}\raggedright
Exception
\end{minipage} & \begin{minipage}[b]{\linewidth}\raggedright
TCGA-only IntOGen candidates
\end{minipage} & \begin{minipage}[b]{\linewidth}\raggedright
Activating MoA (n, \%)
\end{minipage} \\
\midrule\noalign{}
BRCA & 12 & 11/12 & CCND1: conserved\_\allowbreak{}fraction = 0.017 (MYCN) & 39 & 28
(72\%) \\
COAD & 13 & 12/13 & RNF43: conserved\_\allowbreak{}fraction = 0.012 (CASK) & 40 & 34
(85\%) \\
\bottomrule\noalign{}
\end{tabular}
}
\end{minipage}

\emph{Gene universe overlap: 71.4\% of TCGA BRCA genes (13,939/19,514)
and 71.5\% of TCGA COAD genes (14,163/19,795) are present in the GREmLN
epithelial\_cell universe; 95.3\% and 96.9\% of GREmLN genes are
represented in the TCGA BRCA and COAD universes, respectively. BRCA1 has
no regulators in the TCGA BRCA network and is not among the 12
analyzable BRCA focal genes.}

\textbf{Near-zero cross-network overlap}

Regulator set overlap was near-zero across all 25 focal gene--cancer
type pairs (Table 2), so the analysis focuses on the TCGA-only and
GREmLN-only tiers, each evaluated against its own network's gene
universe (Section 2.5). This non-overlap partly reflects coverage: of
the TCGA regulators of the focal genes, 8--9\% are absent from the
GREmLN network and a further 36\% are present but never act as
regulators there, so 44--45\% could not be classified as Both regardless
of biology. The remainder likely reflects differences in regulatory
topology between the two network sources, arising from distinct
inference contexts (single-cell vs.~bulk RNA-seq, pan-tissue atlas
vs.~tumor cohort) and MI threshold calibrations.

\subsubsection{3.2 BRCA: TCGA-Only Candidate
Enrichment}\label{brca-tcga-only-candidate-enrichment}

Across twelve analyzable BRCA focal genes, TCGA-only candidates were
enriched for IntOGen genes against the all-gene background (Stouffer
combined Z = 4.72; Table 3, Figure 3A). Seven of twelve focal genes
reached BH-FDR \textless{} 0.05, with odds ratios from 2.94 (ESR1) to
15.31 (PTEN). TP53's 33 TCGA-only candidates included a single IntOGen
gene (OR = 0.95). Section 3.5 tests this enrichment against the
candidates of random genes, which also show enrichment against the
all-gene background.

\noindent\begin{minipage}{\linewidth}
\textbf{Table 3. BRCA enrichment results (tumor-selective candidates
vs.~IntOGen).}
\par\medskip
{\def\LTcaptype{none}
\begin{tabular}{@{}
  >{\raggedright\arraybackslash}p{(\linewidth - 8\tabcolsep) * \real{0.2000}}
  >{\raggedright\arraybackslash}p{(\linewidth - 8\tabcolsep) * \real{0.2000}}
  >{\raggedright\arraybackslash}p{(\linewidth - 8\tabcolsep) * \real{0.2000}}
  >{\raggedright\arraybackslash}p{(\linewidth - 8\tabcolsep) * \real{0.2000}}
  >{\raggedright\arraybackslash}p{(\linewidth - 8\tabcolsep) * \real{0.2000}}@{}}
\toprule\noalign{}
\begin{minipage}[b]{\linewidth}\raggedright
Focal gene
\end{minipage} & \begin{minipage}[b]{\linewidth}\raggedright
Tumor-selective candidates
\end{minipage} & \begin{minipage}[b]{\linewidth}\raggedright
IntOGen overlap (n)
\end{minipage} & \begin{minipage}[b]{\linewidth}\raggedright
OR
\end{minipage} & \begin{minipage}[b]{\linewidth}\raggedright
BH-FDR
\end{minipage} \\
\midrule\noalign{}
RB1 & 18 & 5 & 11.81 & 0.002** \\
PTEN & 9 & 3 & 15.31 & 0.014* \\
GATA3 & 84 & 8 & 3.24 & 0.021* \\
PIK3CA & 16 & 3 & 7.06 & 0.033* \\
ESR1 & 80 & 7 & 2.94 & 0.033* \\
CCND1 & 18 & 3 & 6.12 & 0.036* \\
BRCA2 & 20 & 3 & 5.40 & 0.042* \\
CTNNB1 & 10 & 2 & 7.64 & 0.057 \\
ERBB2 & 21 & 2 & 3.21 & 0.190 \\
MYC & 16 & 1 & 2.03 & 0.484 \\
CDH1 & 25 & 1 & 1.27 & 0.604 \\
TP53 & 33 & 1 & 0.95 & 0.656 \\
\textbf{Combined} & --- & --- & \textbf{Z=4.72} & --- \\
\bottomrule\noalign{}
\end{tabular}
}
\end{minipage}

*FDR \textless{} 0.05; **FDR \textless{} 0.01 (BH-corrected across focal
genes). Background: TCGA BRCA gene universe (19,514 genes; 620 IntOGen
genes). No combined p-value is given: random genes' candidates are also
enriched against this background, so the test of the panel is the
comparison with random genes (Section 3.5).

MoA directions for candidates of RB1 are shown in Table 4. Two are
consistent with known pathway relationships. CDKN2A carries a repressive
MoA (inverse co-expression with RB1), consistent with the well-described
CDK4/6--RB1 axis, in which pRB represses transcription of CDKN2A
(p16INK4a), so that p16INK4a expression rises when RB1 is lost {[}Li et
al., 1994; Sherr \& Roberts, 1999{]}; note that the known regulation
runs from RB1 to CDKN2A; ARACNe does not infer edge direction, and the
TCGA BRCA network lists this pair in both directions. ARID2, a
chromatin-remodeling tumor suppressor {[}Li et al., 2011{]} reported to
repress cyclin D1 and cyclin E1 {[}Duan et al., 2016{]}, carries an
activating MoA (positive co-expression with RB1), consistent with a
shared role in restraining G1/S progression. These are consistent with,
not validations of, the MoA annotation; systematic validation is beyond
the scope of this work.

\noindent\begin{minipage}{\linewidth}
\textbf{Table 4. RB1 BRCA candidate shortlist (TCGA-only
IntOGen-overlapping regulators).}
\par\medskip
{\def\LTcaptype{none}
\begin{tabular}{@{}lllll@{}}
\toprule\noalign{}
Regulator & Source & IntOGen role & MoA & Direction \\
\midrule\noalign{}
ARID2 & TCGA-only & Tumor suppressor & +1.000 & Activating \\
USP8 & TCGA-only & Oncogene & +1.000 & Activating \\
CDKN2A & TCGA-only & Tumor suppressor & \ensuremath{-}1.000 & Repressive \\
RBM38 & TCGA-only & Tumor suppressor & \ensuremath{-}1.000 & Repressive \\
CNOT3 & TCGA-only & Oncogene & \ensuremath{-}1.000 & Repressive \\
\bottomrule\noalign{}
\end{tabular}
}
\end{minipage}

\subsubsection{3.3 COAD: TCGA-Only Candidate
Enrichment}\label{coad-tcga-only-candidate-enrichment}

In colorectal cancer, TCGA-only candidates were likewise enriched
against the all-gene background (Stouffer Z = 5.88; Table 5, Figure 3B).
Five of thirteen focal genes reached BH-FDR \textless{} 0.05; BRAF
showed the strongest enrichment (5 of 11 candidates are IntOGen genes;
OR = 25.89).

PTEN in COAD had no IntOGen genes among its 18 TCGA-only candidates (OR
= 0.00), so the zero enrichment does not reflect a small candidate set.
In CRC, PTEN is inactivated through mutation {[}Serebriiskii et al.,
2022{]} and through combinations of mutation or 10q23 loss with promoter
hypermethylation {[}Molinari \& Frattini, 2014{]}. Because ARACNe
reconstructs edges from mutual information between expression profiles,
genes inactivated mainly through copy-number loss or epigenetic
silencing may yield less informative regulons; this is a possible
explanation, not a tested one.

\noindent\begin{minipage}{\linewidth}
\textbf{Table 5. COAD enrichment results (tumor-selective candidates
vs.~IntOGen).}
\par\medskip
{\def\LTcaptype{none}
\begin{tabular}{@{}
  >{\raggedright\arraybackslash}p{(\linewidth - 8\tabcolsep) * \real{0.2000}}
  >{\raggedright\arraybackslash}p{(\linewidth - 8\tabcolsep) * \real{0.2000}}
  >{\raggedright\arraybackslash}p{(\linewidth - 8\tabcolsep) * \real{0.2000}}
  >{\raggedright\arraybackslash}p{(\linewidth - 8\tabcolsep) * \real{0.2000}}
  >{\raggedright\arraybackslash}p{(\linewidth - 8\tabcolsep) * \real{0.2000}}@{}}
\toprule\noalign{}
\begin{minipage}[b]{\linewidth}\raggedright
Focal gene
\end{minipage} & \begin{minipage}[b]{\linewidth}\raggedright
Tumor-selective candidates
\end{minipage} & \begin{minipage}[b]{\linewidth}\raggedright
IntOGen overlap (n)
\end{minipage} & \begin{minipage}[b]{\linewidth}\raggedright
OR
\end{minipage} & \begin{minipage}[b]{\linewidth}\raggedright
BH-FDR
\end{minipage} \\
\midrule\noalign{}
BRAF & 11 & 5 & 25.89 & \textless0.001** \\
KRAS & 34 & 6 & 6.66 & 0.004** \\
APC & 21 & 4 & 7.29 & 0.016* \\
FBXW7 & 13 & 3 & 9.29 & 0.023* \\
PIK3CA & 29 & 4 & 4.96 & 0.032* \\
TP53 & 35 & 4 & 4.00 & 0.051 \\
TCF7L2 & 26 & 3 & 4.04 & 0.087 \\
MYC & 16 & 2 & 4.41 & 0.136 \\
RNF43 & 55 & 4 & 2.43 & 0.136 \\
CCND1 & 22 & 2 & 3.09 & 0.192 \\
CTNNB1 & 23 & 2 & 2.94 & 0.192 \\
SMAD4 & 10 & 1 & 3.43 & 0.296 \\
PTEN & 18 & 0 & 0.00 & 1.000 \\
\textbf{Combined} & --- & --- & \textbf{Z=5.88} & --- \\
\bottomrule\noalign{}
\end{tabular}
}
\end{minipage}

*FDR \textless{} 0.05; **FDR \textless{} 0.01 (BH-corrected across focal
genes). Background: TCGA COAD gene universe (19,795 genes; 622 IntOGen
genes). No combined p-value is given (see Table 3 note).

\subsubsection{3.4 GREmLN-Only Candidates Show No Cancer-Gene-Specific
Enrichment}\label{gremln-only-candidates-show-no-cancer-gene-specific-enrichment}

Applying the same per-gene test to GREmLN-only candidate sets, against
the GREmLN epithelial\_cell universe (14,621 genes, 575 IntOGen genes),
gave combined Stouffer Z = 7.83 (BRCA) and 6.94 (COAD) (Tables 6 and 7,
Figure 3C--D). However, this enrichment is not specific to cancer genes.
Random genes' GREmLN-only candidates reach similar values (median
Stouffer Z of random panels of the same size, shown for comparison:
6.1--6.8 for BRCA, 5.6--6.5 for COAD), and the focal genes do not exceed
random genes (Section 3.5). The housekeeping and neutral control genes
also appear ``enriched'' in this tier (BH-FDR \textless{} 0.05 for 4 of
5 housekeeping and 4 of 4 neutral genes, in both cancer types). The
GREmLN-only per-gene results therefore reflect the general enrichment of
regulators for driver genes, not cancer-gene-specific regulation, and
should not be interpreted as evidence for individual focal genes.

GREmLN-only candidates are a focal gene's GREmLN regulators minus any
that are also its TCGA regulators in that cancer type. Because a focal
gene's GREmLN and TCGA regulators almost never coincide (Section 3.1),
these sets are nearly identical in BRCA and COAD: among the six focal
genes in both panels (TP53, MYC, CTNNB1, CCND1, PIK3CA, PTEN), the only
difference is CCND1 (41 vs 42 candidates): in BRCA, its GREmLN regulator
MYCN is also one of its TCGA regulators, so MYCN is labelled Both rather
than GREmLN-only.

\noindent\begin{minipage}{\linewidth}
\textbf{Table 6. GREmLN-only enrichment results (BRCA; GREmLN
epithelial\_cell background).}
\par\medskip
{\def\LTcaptype{none}
\begin{tabular}{@{}
  >{\raggedright\arraybackslash}p{(\linewidth - 8\tabcolsep) * \real{0.2000}}
  >{\raggedright\arraybackslash}p{(\linewidth - 8\tabcolsep) * \real{0.2000}}
  >{\raggedright\arraybackslash}p{(\linewidth - 8\tabcolsep) * \real{0.2000}}
  >{\raggedright\arraybackslash}p{(\linewidth - 8\tabcolsep) * \real{0.2000}}
  >{\raggedright\arraybackslash}p{(\linewidth - 8\tabcolsep) * \real{0.2000}}@{}}
\toprule\noalign{}
\begin{minipage}[b]{\linewidth}\raggedright
Focal gene
\end{minipage} & \begin{minipage}[b]{\linewidth}\raggedright
GREmLN-only candidates
\end{minipage} & \begin{minipage}[b]{\linewidth}\raggedright
IntOGen overlap (n)
\end{minipage} & \begin{minipage}[b]{\linewidth}\raggedright
OR
\end{minipage} & \begin{minipage}[b]{\linewidth}\raggedright
BH-FDR
\end{minipage} \\
\midrule\noalign{}
RB1 & 17 & 8 & 22.01 & \textless0.001** \\
MYC & 25 & 8 & 11.64 & \textless0.001** \\
ERBB2 & 48 & 13 & 9.26 & \textless0.001** \\
CTNNB1 & 18 & 4 & 7.02 & 0.011* \\
CDH1 & 18 & 4 & 7.02 & 0.011* \\
TP53 & 7 & 2 & 9.80 & 0.049* \\
ESR1 & 7 & 2 & 9.80 & 0.049* \\
GATA3 & 24 & 3 & 3.50 & 0.100 \\
BRCA2 & 20 & 2 & 2.72 & 0.246 \\
CCND1 & 41 & 3 & 1.93 & 0.261 \\
BRCA1 & 23 & 1 & 1.11 & 0.658 \\
PTEN & 4 & 0 & 0.00 & 1.000 \\
\textbf{Combined} & --- & --- & \textbf{Z=7.83} & --- \\
\bottomrule\noalign{}
\end{tabular}
}
\end{minipage}

*FDR \textless{} 0.05; **FDR \textless{} 0.01 (BH-corrected across
testable focal genes). PIK3CA excluded (n/a in Figure 3C): only 2
GREmLN-only candidates (n \textless{} 3, not tested). BRCA1 has no TCGA
BRCA regulators (not in Table 3) but is testable here. No combined
p-value is given: the comparison with random genes (Section 3.5) shows
this Z is not specific to cancer genes.

\noindent\begin{minipage}{\linewidth}
\textbf{Table 7. GREmLN-only enrichment results (COAD; GREmLN
epithelial\_cell background).}
\par\medskip
{\def\LTcaptype{none}
\begin{tabular}{@{}
  >{\raggedright\arraybackslash}p{(\linewidth - 8\tabcolsep) * \real{0.2000}}
  >{\raggedright\arraybackslash}p{(\linewidth - 8\tabcolsep) * \real{0.2000}}
  >{\raggedright\arraybackslash}p{(\linewidth - 8\tabcolsep) * \real{0.2000}}
  >{\raggedright\arraybackslash}p{(\linewidth - 8\tabcolsep) * \real{0.2000}}
  >{\raggedright\arraybackslash}p{(\linewidth - 8\tabcolsep) * \real{0.2000}}@{}}
\toprule\noalign{}
\begin{minipage}[b]{\linewidth}\raggedright
Focal gene
\end{minipage} & \begin{minipage}[b]{\linewidth}\raggedright
GREmLN-only candidates
\end{minipage} & \begin{minipage}[b]{\linewidth}\raggedright
IntOGen overlap (n)
\end{minipage} & \begin{minipage}[b]{\linewidth}\raggedright
OR
\end{minipage} & \begin{minipage}[b]{\linewidth}\raggedright
BH-FDR
\end{minipage} \\
\midrule\noalign{}
MYC & 25 & 8 & 11.64 & \textless0.001** \\
APC & 23 & 6 & 8.70 & 0.001** \\
FBXW7 & 21 & 5 & 7.69 & 0.004** \\
CTNNB1 & 18 & 4 & 7.02 & 0.011* \\
SMAD4 & 27 & 5 & 5.59 & 0.011* \\
TP53 & 7 & 2 & 9.80 & 0.057 \\
BRAF & 32 & 4 & 3.51 & 0.061 \\
CCND1 & 42 & 4 & 2.58 & 0.123 \\
TCF7L2 & 14 & 2 & 4.08 & 0.135 \\
RNF43 & 30 & 3 & 2.72 & 0.135 \\
KRAS & 7 & 1 & 4.08 & 0.267 \\
PTEN & 4 & 0 & 0.00 & 1.000 \\
\textbf{Combined} & --- & --- & \textbf{Z=6.94} & --- \\
\bottomrule\noalign{}
\end{tabular}
}
\end{minipage}

*FDR \textless{} 0.05; **FDR \textless{} 0.01 (BH-corrected across
testable focal genes). PIK3CA excluded (n/a in Figure 3D): only 2
GREmLN-only candidates (n \textless{} 3, not tested). No combined
p-value is given (see Table 6 note).

\subsubsection{3.5 Comparison with Random Genes and Negative
Controls}\label{comparison-with-random-genes-and-negative-controls}

\textbf{Random genes.} Because regulators are about 2.7-fold richer in
IntOGen genes than genes overall (Section 2.5), random genes' candidates
also show enrichment against the all-gene background: for TCGA-only
candidates, random panels of the same size as the focal panels reach a
median Stouffer Z of 3.2--3.4 (BRCA) and 3.8--4.4 (COAD) (shown for
comparison, not as a test), compared with the focal panels' 4.72 and
5.88. Comparing effect sizes directly (Table S3):

\begin{itemize}
\tightlist
\item
  \textbf{TCGA-only, COAD:} the focal genes' candidates exceed those of
  random genes (geometric-mean OR 5.04 vs 3.21--3.31; nominal
  permutation p = 0.012--0.036 across the five random samples).
\item
  \textbf{TCGA-only, BRCA:} the difference is borderline (geometric-mean
  OR 4.92 vs 3.34--3.48; p = 0.048--0.083).
\item
  \textbf{GREmLN-only:} no difference from random genes in either cancer
  type (BRCA p = 0.229--0.319; COAD p = 0.209--0.467).
\end{itemize}

With a regulator-only background, against which random genes show much
weaker enrichment, the conclusions are unchanged (TCGA-only: COAD p =
0.012--0.037, BRCA p = 0.049--0.083; GREmLN-only: p \ensuremath{\geq} 0.20).

\textbf{Housekeeping and neutral controls.} In the TCGA-only tier, no
housekeeping or neutral control gene reached BH-FDR \textless{} 0.05 in
either cancer type (Tables 8 and 9; Figures 4 and 5); the single
nominally significant result was LDHA in COAD (OR = 4.42, p = 0.038),
which did not replicate in BRCA (OR = 0.00). The control panels'
geometric-mean odds ratios, computed as for random genes, were below
those of random genes (1.4--2.5 vs 3.2--3.5). In the GREmLN-only tier,
by contrast, most control genes appeared enriched (Section 3.4).

Taken together, TCGA-only candidates of cancer driver genes are modestly
enriched for known driver genes beyond the general enrichment of
regulators --- in COAD (p \textless{} 0.05 without multiple-testing
correction in all five random samples; see Limitations) and borderline
in BRCA --- while GREmLN-only candidates are not.

\noindent\begin{minipage}{\linewidth}
\textbf{Table 8. Housekeeping control results: housekeeping gene
tumor-selective candidates vs.~IntOGen.}
\par\medskip
{\def\LTcaptype{none}
\begin{tabular}{@{}lllll@{}}
\toprule\noalign{}
Gene & Context & Tumor-selective candidates & OR & p-value \\
\midrule\noalign{}
ACTB & BRCA & 42 & 0.00 & 1.000 \\
GAPDH & BRCA & 16 & 2.03 & 0.404 \\
HPRT1 & BRCA & 29 & 2.26 & 0.235 \\
LDHA & BRCA & 27 & 0.00 & 1.000 \\
TUBB & BRCA & 16 & 2.03 & 0.404 \\
ACTB & COAD & 18 & 1.81 & 0.437 \\
GAPDH & COAD & 23 & 1.40 & 0.520 \\
HPRT1 & COAD & 10 & 0.00 & 1.000 \\
LDHA & COAD & 24 & 4.42 & 0.038 \\
TUBB & COAD & 15 & 0.00 & 1.000 \\
\bottomrule\noalign{}
\end{tabular}
}
\end{minipage}

\noindent\begin{minipage}{\linewidth}
\textbf{Table 9. Neutral control results: tumor-expressed non-driver
gene tumor-selective candidates vs.~IntOGen.}
\par\medskip
{\def\LTcaptype{none}
\begin{tabular}{@{}lllll@{}}
\toprule\noalign{}
Gene & Context & Tumor-selective candidates & OR & p-value \\
\midrule\noalign{}
FASN & BRCA & 8 & 0.00 & 1.000 \\
PCNA & BRCA & 17 & 1.91 & 0.423 \\
PKM & BRCA & 24 & 1.33 & 0.539 \\
VIM & BRCA & 22 & 3.05 & 0.154 \\
FASN & COAD & 30 & 2.21 & 0.243 \\
PCNA & COAD & 19 & 1.71 & 0.455 \\
PKM & COAD & 30 & 2.21 & 0.243 \\
VIM & COAD & 39 & 0.81 & 0.712 \\
\bottomrule\noalign{}
\end{tabular}
}
\end{minipage}

\subsubsection{3.6 Candidate Profiling}\label{candidate-profiling}

Shortlisted candidates can be passed to
\texttt{comprehensive\_gene\_analysis}, which profiles each gene with
the rule-based domain assessments of Section 2.1 (oncogenic potential,
druggability, clinical actionability, network vulnerability) together
with PageRank centrality, in either a TCGA or a GREmLN network context.
These assessments are coarse topology-derived prioritization signals:
highly connected regulators tend to receive the highest rating in every
domain. This operation is illustrative and is not evaluated in this
paper.

\begin{center}\rule{0.5\linewidth}{0.5pt}\end{center}

\subsection{4. Discussion}\label{discussion}

We have presented RegNetAgents --- to our knowledge, the first framework
to unify dual-network ARACNe source classification, cancer-driver
filtering, and MoA annotation in a single reproducible workflow --- and
evaluated whether its source-labeled candidate lists for cancer driver
genes are enriched for known driver genes. Each candidate is annotated
with its network source and, where available, MoA directionality,
turning an undifferentiated regulator list into an ordered shortlist
with directional context.

\subsubsection{Interpreting the
enrichment}\label{interpreting-the-enrichment}

Because regulator sets are themselves enriched for driver genes (about
8.5\% of TCGA regulators are IntOGen genes, compared with 3.1--3.2\% of
all genes), enrichment measured against an all-gene background mixes two
components: the general driver-richness of regulators, which any gene's
regulator set shares, and any additional signal specific to the queried
cancer gene. The comparison with random genes was designed to separate
the two. After this comparison, TCGA-only candidates retain a modest
cancer-gene-related excess (in COAD, p \textless{} 0.05 without
multiple-testing correction in all five random samples; borderline in
BRCA), while GREmLN-only candidates retain none. The claim supported
here is limited accordingly: the TCGA-only regulators of cancer driver
genes are somewhat more often driver genes than those of random
non-driver genes (in COAD; borderline in BRCA). It is not a claim that
individual candidates are drivers of the focal gene, nor that the signal
is specific to the particular driver genes chosen.

That the excess is confined to the TCGA-only tier is consistent with the
TCGA networks being cancer-type-matched tumor networks, whereas the
GREmLN epithelial\_cell network is the pan-tissue population average
described in Section 2.3. It is equally consistent with technical
differences between the sources --- single-cell versus bulk inference,
and a roughly three-fold smaller regulator set in GREmLN --- and with
IntOGen and the TCGA networks drawing partly on the same patient cohorts
(mutation versus expression data). These explanations cannot be
separated with the present data.

\subsubsection{Practical implications}\label{practical-implications}

For a user, the source labels and MoA annotations are the main
deliverables: they reduce a focal gene's regulatory neighborhood to a
short, annotated list. The enrichment results indicate that TCGA-only
candidates of cancer genes carry a modest excess of known driver genes,
so the lists are better treated as hypothesis-generating than as
confirmatory. GREmLN-only candidates, which the tool also reports,
showed no excess over random genes; their label indicates network
context (found in the GREmLN reference but not the tumor network), not
additional evidence of cancer relevance.

\subsubsection{Limitations}\label{limitations}

Both network sources represent population averages --- GREmLN averaging
across single cells within a cell type, TCGA across bulk tumors --- and
neither captures cell-state dynamics or tumor subclone heterogeneity.
The \textasciitilde3\ensuremath{\times} regulator count difference between backends biases
classification toward TCGA-only, and 44--45\% of the focal genes' TCGA
regulators could not appear as GREmLN regulators (Section 3.1); the
comparison with random genes accounts for the general enrichment of
regulators but not for every structural difference between the networks.

The enrichment effects are modest. In COAD the TCGA-only excess over
random genes has p \textless{} 0.05 without multiple-testing correction
in all five random samples; in BRCA it is borderline (p = 0.048--0.083).
Several focal genes show no individual enrichment (e.g., TP53 and CDH1
in BRCA, PTEN in COAD). The combined Stouffer Z values (Tables 3 and
5--7) are descriptive only. Panels of random genes also score high on
this measure, so a high Z is not evidence on its own; the comparison
with random genes (Section 3.5) is the actual test.

The comparison with random genes has further statistical limitations. No
correction across the four comparisons (two cancer types \ensuremath{\times} two tiers)
was pre-specified; with Bonferroni correction (p \textless{} 0.0125),
the COAD excess remains significant in one of the five random samples.
With 12--13 focal genes per cancer type, the tests have limited power.
The permutation test treats genes as exchangeable, although genes that
share regulators have correlated odds ratios, which may make the
p-values somewhat optimistic. Finally, random genes were drawn from
non-IntOGen genes whereas the focal genes are IntOGen drivers, so the
comparison does not separate properties of the chosen focal genes from
properties of driver genes in general; the focal genes are also well
studied and may differ from random genes in network connectivity.

IntOGen identifies drivers from somatic mutation data that partly
include TCGA cohorts, whereas the TCGA networks are inferred from
expression data of TCGA tumors. The data types differ, but the shared
cohorts mean the reference set and the networks are not fully
independent.

We used a single reference cell type (epithelial cell), the only GREmLN
cell type with sufficient gene coverage for solid-tumor comparisons. The
epithelial\_cell network is a pan-tissue average derived from the
CELLxGENE Census --- aggregating diverse healthy, disease-adjacent, and
pathological tissue sources --- and does not represent a strictly normal
or tissue-specific baseline; tissue-specific regulatory relationships
may be diluted.

MoA annotations reflect the direction of expression correlation, not
demonstrated regulatory mechanisms; the examples discussed are
consistent with known pathway relationships but do not constitute
systematic validation. The rule-based assessments of
\texttt{comprehensive\_gene\_analysis} are derived from network topology
and were not evaluated. No empirical comparison to VIPER, PANDA, or
SCENIC was performed; the tool comparisons in the Introduction are
conceptual. These tools address different questions (VIPER infers
regulator activity from expression signatures; PANDA and SCENIC infer
networks), so a direct benchmark would first require defining a common
task and ground truth, which is beyond the scope of this work.

\subsubsection{Future directions}\label{future-directions}

RegNetAgents' architecture supports extension to additional network
types. Immune cell network comparisons (contrasting immune-cell networks
inferred from tumor samples with GREmLN project immune cell ARACNe
networks) would extend the framework to tumor microenvironment analysis.
A multi-cancer panel analysis across all fourteen TCGA cancer types,
with the random-gene comparison introduced here, would test whether the
TCGA-only signal generalizes, and comparing focal genes with other
driver genes as well as with random genes would test whether it is
specific to the chosen focal genes. Because the GREmLN epithelial\_cell
network is a pan-tissue average rather than a tissue-matched one
(Limitations), a further comparison would contrast the TCGA tumor
networks with tissue-matched single-cell ARACNe networks (for example, a
breast-tissue network against TCGA BRCA), where such networks are
available; this would help show whether the absence of a GREmLN-only
excess here reflects the pan-tissue baseline rather than the single-cell
source. Support for user-supplied ARACNe networks would allow the
framework to be applied to custom datasets.

\begin{center}\rule{0.5\linewidth}{0.5pt}\end{center}

\subsection{Funding}\label{funding}

No funding was received for this work.

\subsection{Competing Interests}\label{competing-interests}

The author is the founder of Bird AI Solutions. This work was conducted
independently and received no funding.

\subsection{Acknowledgements}\label{acknowledgements}

Network data were obtained from the GREmLN team's pre-computed ARACNe
networks {[}Zhang et al., 2025{]}. The results shown here are in whole
or part based upon data generated by the TCGA Research Network:
\url{https://www.cancer.gov/tcga}. The author designed the system
architecture, workflow logic, and analysis pipeline, and directed the
statistical analysis.

\subsection{AI Usage Disclosure}\label{ai-usage-disclosure}

Claude Code (Anthropic) was used to assist with software implementation,
statistical analysis, and paper drafting. The author reviewed and
approved all analyses and text and takes full responsibility for the
methods and conclusions.

\subsection{Data Availability}\label{data-availability}

RegNetAgents is available at \url{https://github.com/jab57/RegNetAgents} under
the MIT License. The v1.3.0 release used for this paper is archived on
Zenodo (DOI: 10.5281/zenodo.23215310); the latest version is available
via the concept DOI 10.5281/zenodo.18500027. GREmLN project ARACNe
network files were obtained from the GREmLN Quickstart Tutorial (Chan
Zuckerberg Initiative Virtual Cells Platform;
\url{https://virtualcellmodels.cziscience.com/quickstart/gremln-quickstart};
model code: \url{https://github.com/czi-ai/GREmLN}) under the Virtual Cells
Platform Terms of Use (scientific research use) and are bundled with the
RegNetAgents package for convenience; they derive from CELLxGENE Census
data {[}CZI Cell Science Program et al., 2025{]} (CC BY 4.0,
\url{https://creativecommons.org/licenses/by/4.0/}). No separate data DOI
exists for these files. TCGA ARACNe networks are those of Bioconductor
\texttt{aracne.networks} v1.36.0, whose network files the package
authors publish on Zenodo (doi:10.5281/zenodo.22918956; CC BY-NC-ND 4.0,
\url{https://creativecommons.org/licenses/by-nc-nd/4.0/}; the 14 networks used
here are byte-identical to v1.36.0; RegNetAgents converts them locally
to CSV/PKL format without changing their content); they are not
redistributed with RegNetAgents, and
\path{scripts/setup_tcga_networks.py} installs them from that
record. The IntOGen driver compendium (release 2024.09.20, CC0 1.0,
\url{https://creativecommons.org/publicdomain/zero/1.0/}) was obtained from
\url{https://www.intogen.org} and is bundled with the package. Analysis
scripts are available in the repository under \path{scripts/}. Full
source-labeled lists of IntOGen-overlapping candidates for the
analyzable BRCA and COAD focal genes (network source, IntOGen role, and
MoA annotation) are provided as
\path{supplementary/table_s1_brca_candidates.csv} and
\path{supplementary/table_s2_coad_candidates.csv}, and the
focal-versus-random-gene comparison as
\path{supplementary/table_s3_tier_specificity.csv}. Network-derived
values (Tables 1 and 4; Supplementary Tables S1--S2) remain subject to
the CC BY-NC-ND 4.0 license of the TCGA network data and are not covered
by this article's license.

\subsection{Figure Legends}\label{figure-legends}

\begin{figure}
\centering
\pandocbounded{\includegraphics[keepaspectratio,alt={Figure 1. RegNetAgents workflow for source-labeled candidate regulatory list generation. Schematic of the analysis pipeline. For each focal gene, compare\_network\_contexts queries the GREmLN epithelial cell network and the TCGA ARACNe tumor network, classifies regulators by network source (TCGA-only / GREmLN-only / Both), and filters the result against IntOGen to produce a source-labeled candidate list. Each entry is annotated with its IntOGen role and, for TCGA-sourced entries, MoA direction (activating or repressive). Enrichment is tested per tier with Fisher's exact test, BH-FDR correction, and Stouffer Z, and each tier is compared with the candidates of random genes.}]{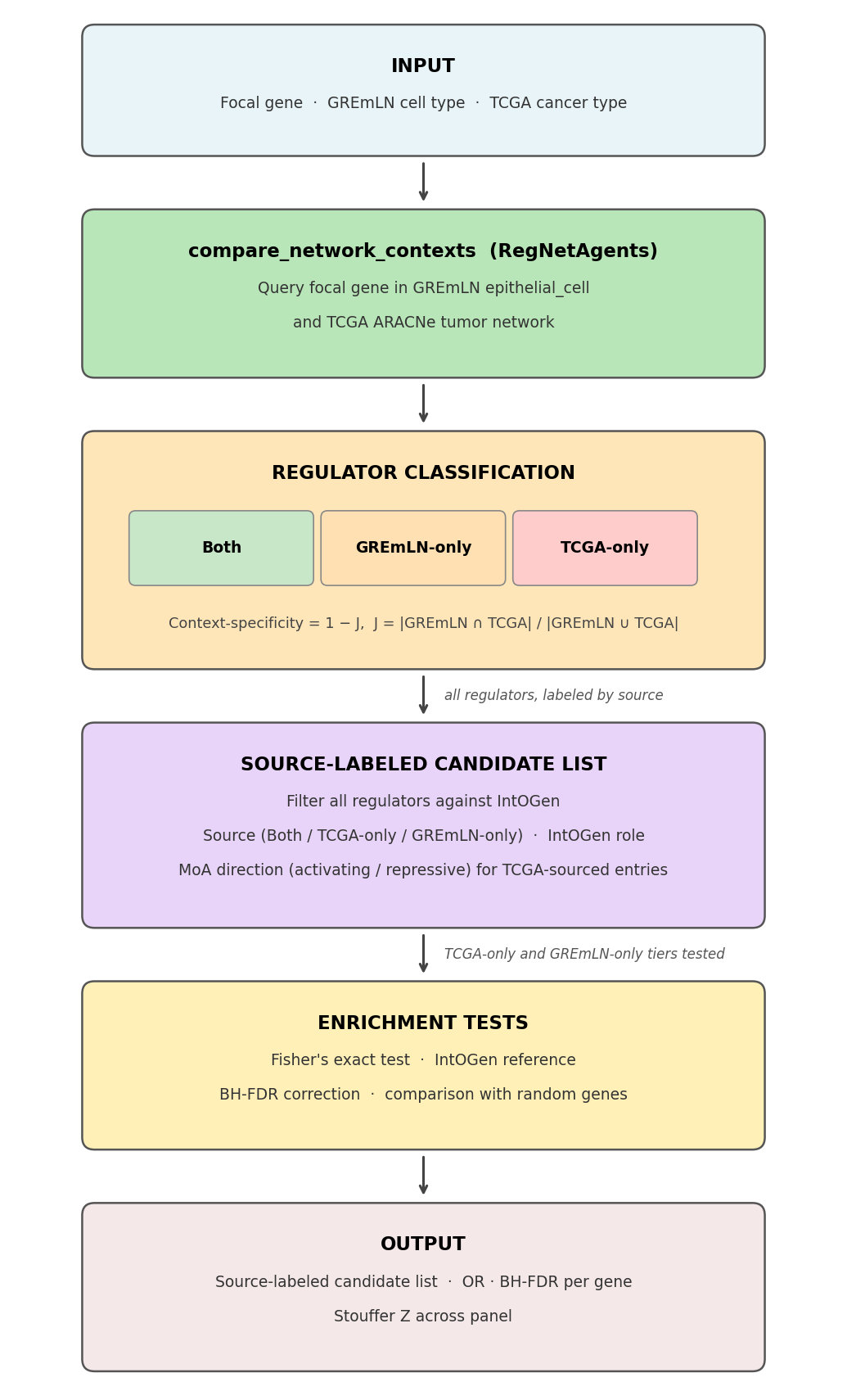}}
\caption{\textbf{Figure 1. RegNetAgents workflow for source-labeled
candidate regulatory list generation.} Schematic of the analysis
pipeline. For each focal gene,
\protect\texttt{compare\_network\_contexts} queries the GREmLN
epithelial cell network and the TCGA ARACNe tumor network, classifies
regulators by network source (TCGA-only / GREmLN-only / Both), and
filters the result against IntOGen to produce a source-labeled candidate
list. Each entry is annotated with its IntOGen role and, for
TCGA-sourced entries, MoA direction (activating or repressive).
Enrichment is tested per tier with Fisher's exact test, BH-FDR
correction, and Stouffer Z, and each tier is compared with the
candidates of random genes.}
\end{figure}

\begin{figure}
\centering
\pandocbounded{\includegraphics[keepaspectratio,alt={Figure 2. Source-labeled IntOGen candidate regulatory lists for BRCA and COAD focal genes. Stacked bar charts showing the number of IntOGen-overlapping regulators per focal gene classified by network source: Both (present in GREmLN epithelial and TCGA tumor networks), TCGA-only (tumor-selective), and GREmLN-only (single-cell ARACNe). (A) Breast cancer (BRCA); twelve focal genes (BRCA1, which has no TCGA regulators, is not shown). MoA annotations of TCGA-sourced candidates are not drawn and are given in Tables S1--S2. This figure illustrates the primary deliverable of each compare\_network\_contexts query.}]{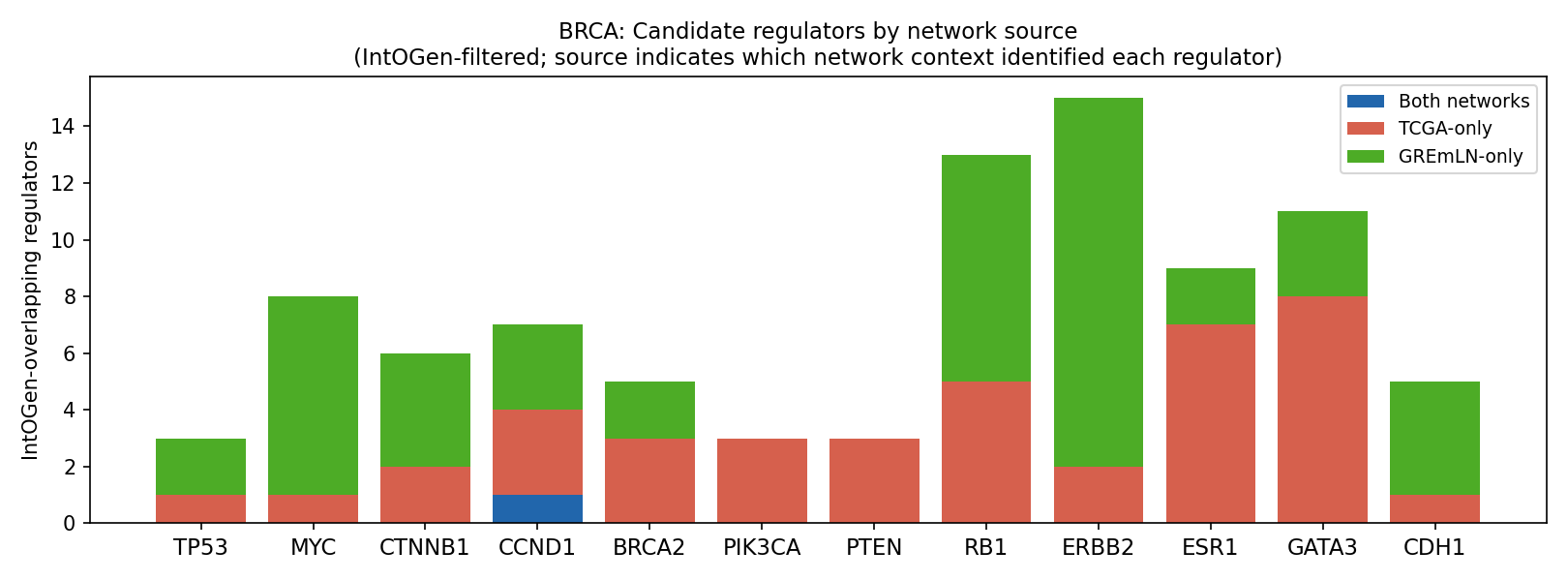}}
\caption{\textbf{Figure 2. Source-labeled IntOGen candidate regulatory
lists for BRCA and COAD focal genes.} Stacked bar charts showing the
number of IntOGen-overlapping regulators per focal gene classified by
network source: Both (present in GREmLN epithelial and TCGA tumor
networks), TCGA-only (tumor-selective), and GREmLN-only (single-cell
ARACNe). \textbf{(A)} Breast cancer (BRCA); twelve focal genes (BRCA1,
which has no TCGA regulators, is not shown). MoA annotations of
TCGA-sourced candidates are not drawn and are given in Tables S1--S2.
This figure illustrates the primary deliverable of each
\protect\texttt{compare\_network\_contexts} query.}
\end{figure}

\begin{figure}
\centering
\pandocbounded{\includegraphics[keepaspectratio,alt={(B) Colorectal cancer (COAD); twelve of thirteen focal genes (PTEN has no IntOGen-overlapping regulators in either network and is not shown).}]{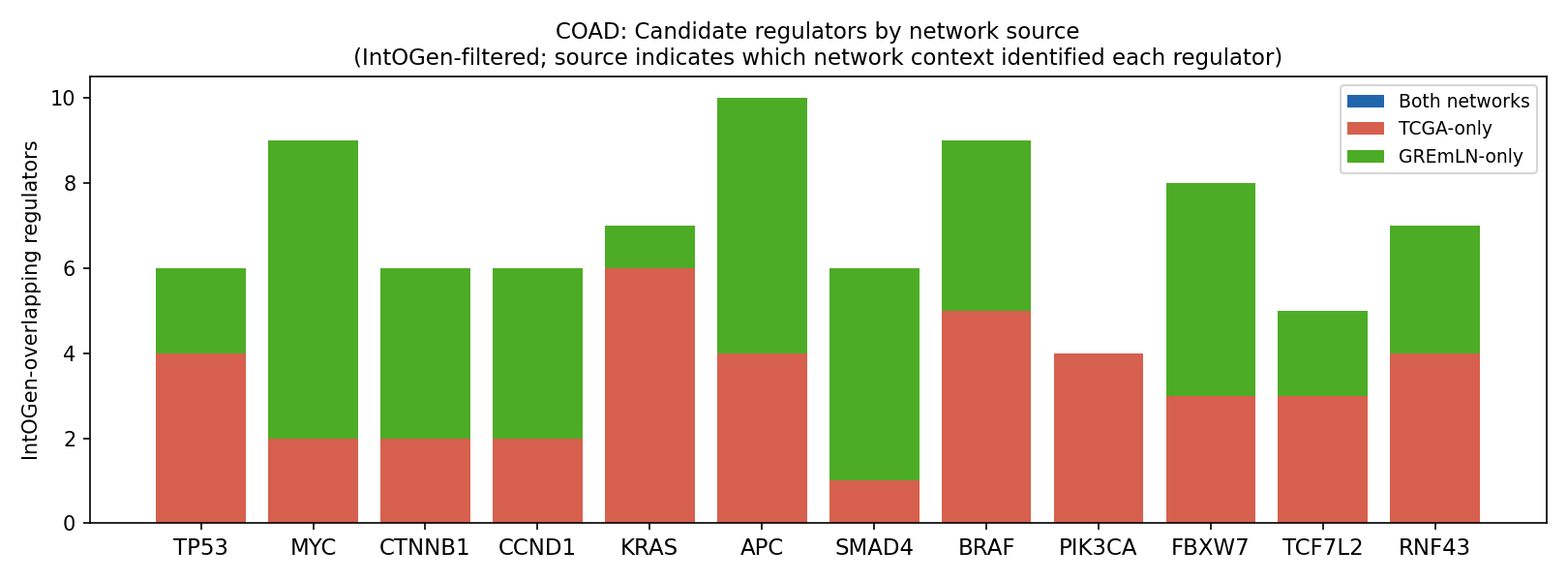}}
\caption{\textbf{(B)} Colorectal cancer (COAD); twelve of thirteen focal
genes (PTEN has no IntOGen-overlapping regulators in either network and
is not shown).}
\end{figure}

\begin{figure}
\centering
\pandocbounded{\includegraphics[keepaspectratio,alt={Figure 3. Per-gene enrichment of candidates for IntOGen driver genes, by network tier. Heatmap of odds ratios (Fisher's exact test, one-tailed) for each focal gene (rows). Cell annotations show the odds ratio; asterisks indicate BH-FDR significance (* \textless{} 0.05, ** \textless{} 0.01). Color scale: blue = enriched (OR \textgreater{} 1), red = depleted (OR \textless{} 1), white = OR \ensuremath{\approx} 1; grey cells labeled n/a = insufficient candidates (n \textless{} 3, not tested; PIK3CA in panels C and D). These are tests against all-gene backgrounds; Section 3.5 compares them with random genes, against which only the TCGA-only tier retains a modest excess (in COAD; borderline in BRCA). (A) BRCA, TCGA-only candidates (background = TCGA BRCA gene universe, 620 IntOGen genes).}]{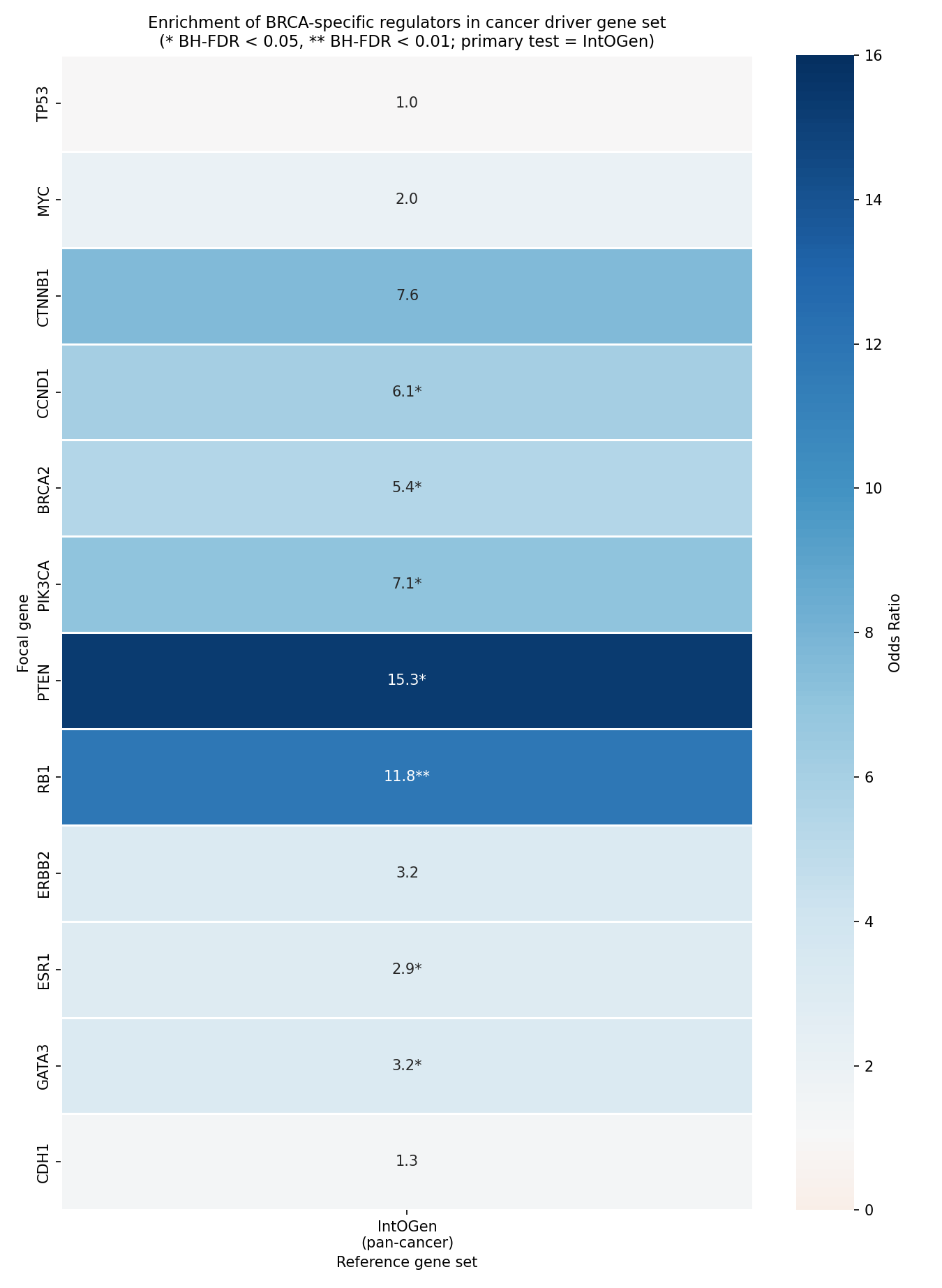}}
\caption{\textbf{Figure 3. Per-gene enrichment of candidates for IntOGen
driver genes, by network tier.} Heatmap of odds ratios (Fisher's exact
test, one-tailed) for each focal gene (rows). Cell annotations show the
odds ratio; asterisks indicate BH-FDR significance (* \textless{} 0.05,
** \textless{} 0.01). Color scale: blue = enriched (OR \textgreater{}
1), red = depleted (OR \textless{} 1), white = OR \ensuremath{\approx} 1; grey cells
labeled n/a = insufficient candidates (n \textless{} 3, not tested;
PIK3CA in panels C and D). These are tests against all-gene backgrounds;
Section 3.5 compares them with random genes, against which only the
TCGA-only tier retains a modest excess (in COAD; borderline in BRCA).
\textbf{(A)} BRCA, TCGA-only candidates (background = TCGA BRCA gene
universe, 620 IntOGen genes).}
\end{figure}

\begin{figure}
\centering
\pandocbounded{\includegraphics[keepaspectratio,alt={(B) COAD, TCGA-only candidates (background = TCGA COAD gene universe, 622 IntOGen genes).}]{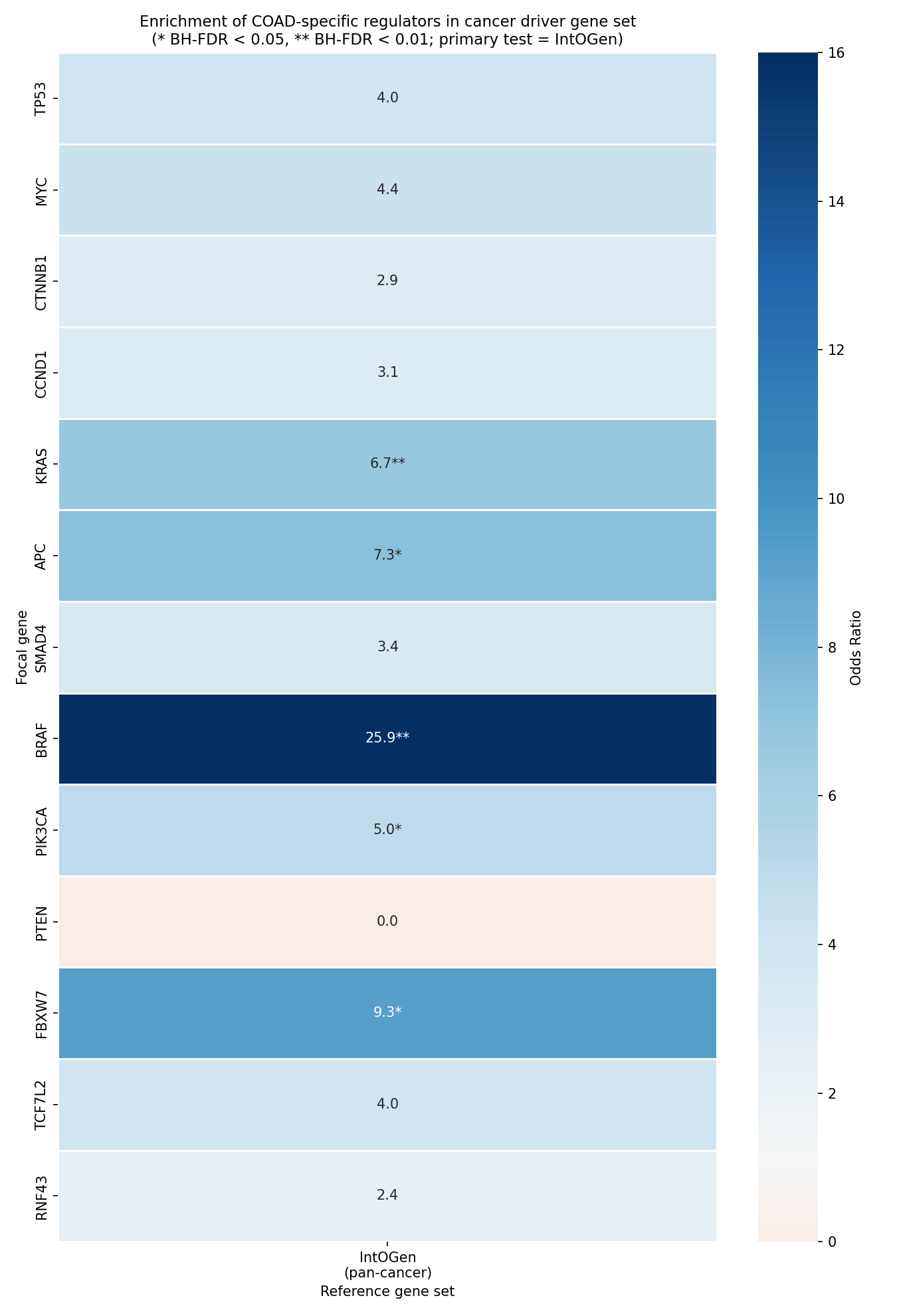}}
\caption{\textbf{(B)} COAD, TCGA-only candidates (background = TCGA COAD
gene universe, 622 IntOGen genes).}
\end{figure}

\begin{figure}
\centering
\pandocbounded{\includegraphics[keepaspectratio,alt={(C) BRCA, GREmLN-only candidates (single-cell ARACNe; background = GREmLN epithelial\_cell universe, 14,621 genes, 575 IntOGen genes).}]{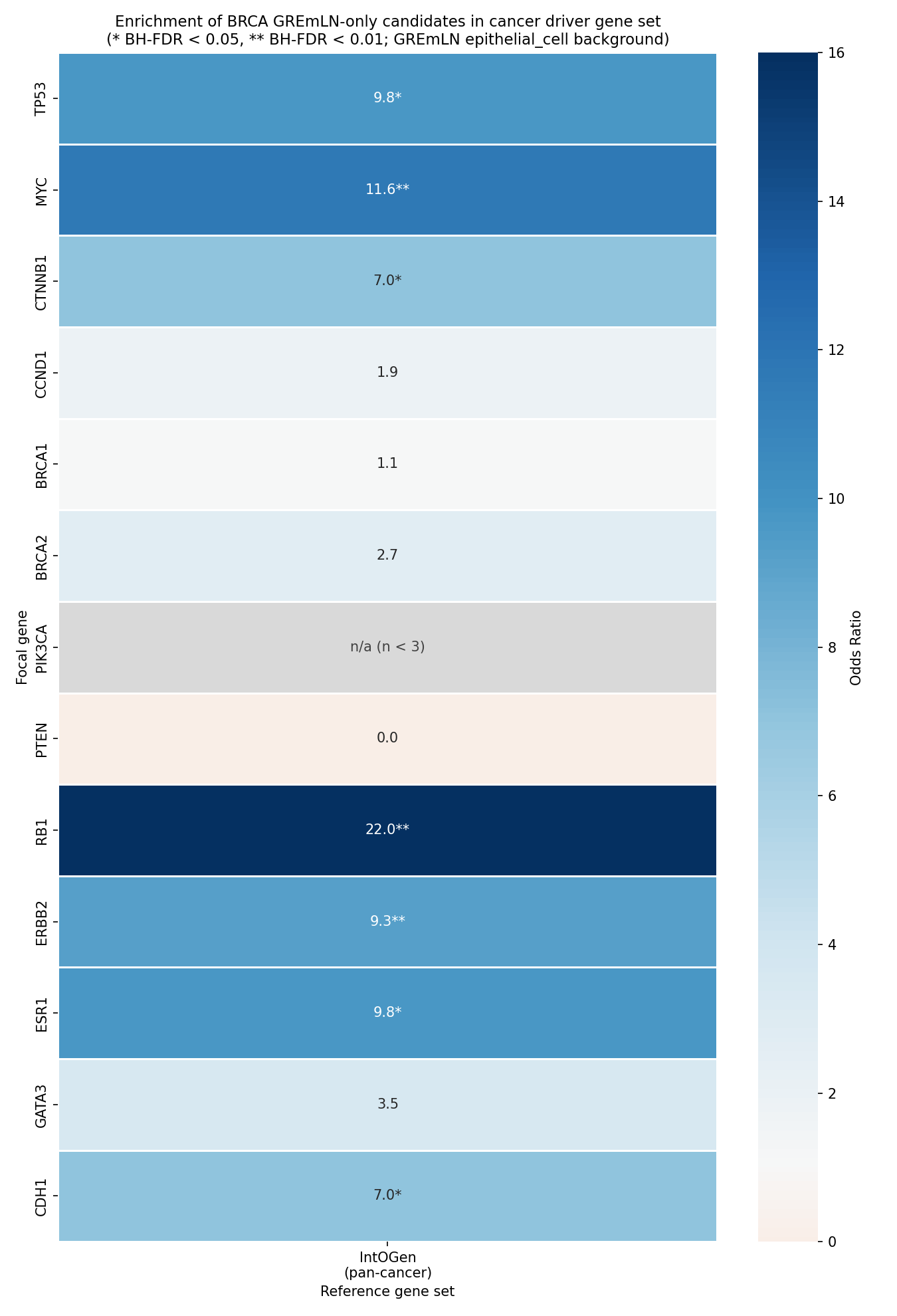}}
\caption{\textbf{(C)} BRCA, GREmLN-only candidates (single-cell ARACNe;
background = GREmLN epithelial\_cell universe, 14,621 genes, 575 IntOGen
genes).}
\end{figure}

\begin{figure}
\centering
\pandocbounded{\includegraphics[keepaspectratio,alt={(D) COAD, GREmLN-only candidates (same GREmLN background).}]{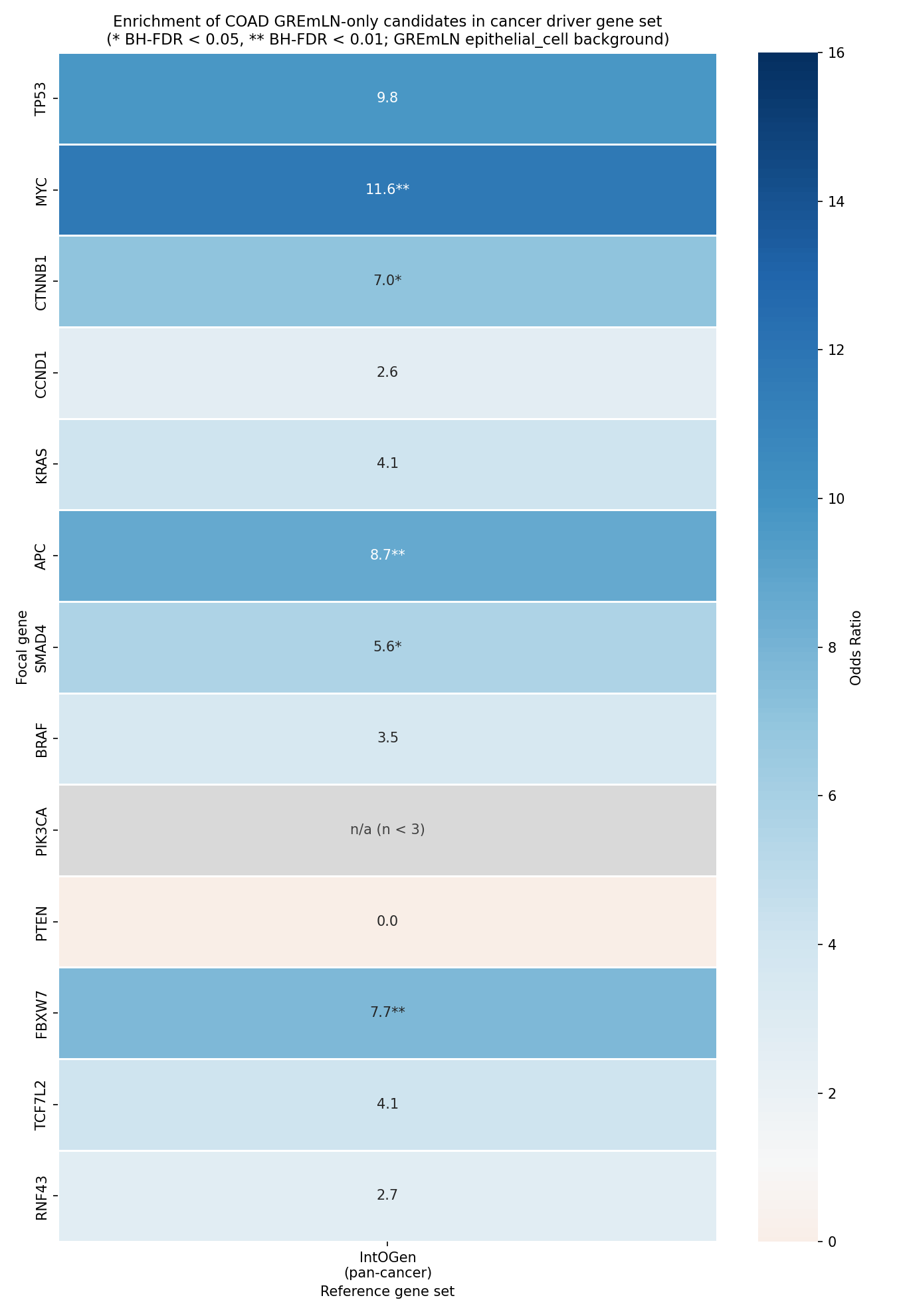}}
\caption{\textbf{(D)} COAD, GREmLN-only candidates (same GREmLN
background).}
\end{figure}

\begin{figure}
\centering
\pandocbounded{\includegraphics[keepaspectratio,alt={Figure 4. Housekeeping gene controls vs.~cancer focal genes. Odds ratios for TCGA-only candidates of five housekeeping genes (ACTB, GAPDH, HPRT1, LDHA, TUBB; teal) compared with the cancer focal genes (red). One control result is nominally significant (LDHA in COAD, OR = 4.42, p = 0.038) and does not replicate in BRCA. Bars above OR = 20 are truncated and labeled with their value (BRAF in COAD, 25.9). (A) BRCA.}]{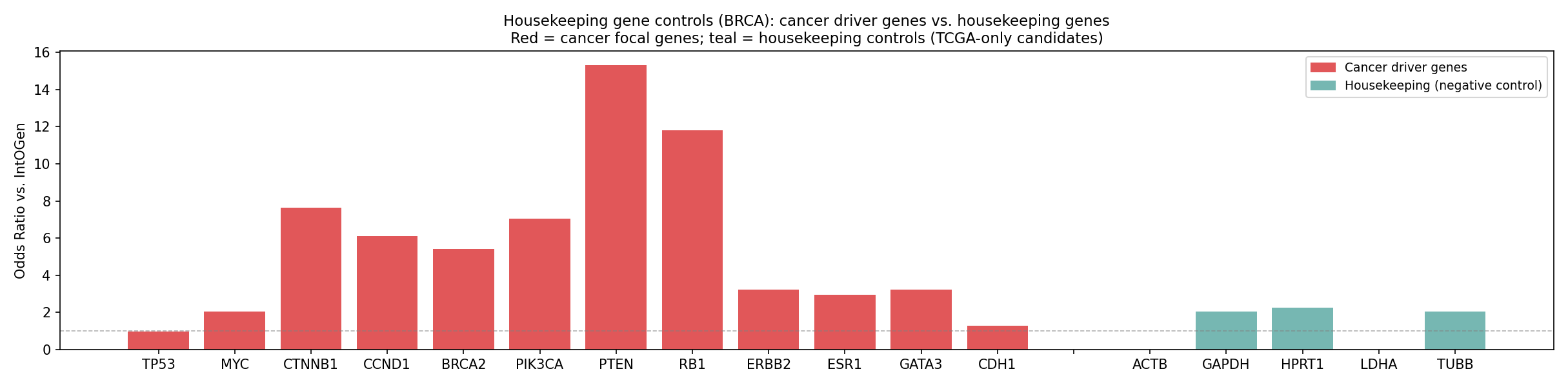}}
\caption{\textbf{Figure 4. Housekeeping gene controls vs.~cancer focal
genes.} Odds ratios for TCGA-only candidates of five housekeeping genes
(ACTB, GAPDH, HPRT1, LDHA, TUBB; teal) compared with the cancer focal
genes (red). One control result is nominally significant (LDHA in COAD,
OR = 4.42, p = 0.038) and does not replicate in BRCA. Bars above OR = 20
are truncated and labeled with their value (BRAF in COAD, 25.9).
\textbf{(A)} BRCA.}
\end{figure}

\begin{figure}
\centering
\pandocbounded{\includegraphics[keepaspectratio,alt={(B) COAD.}]{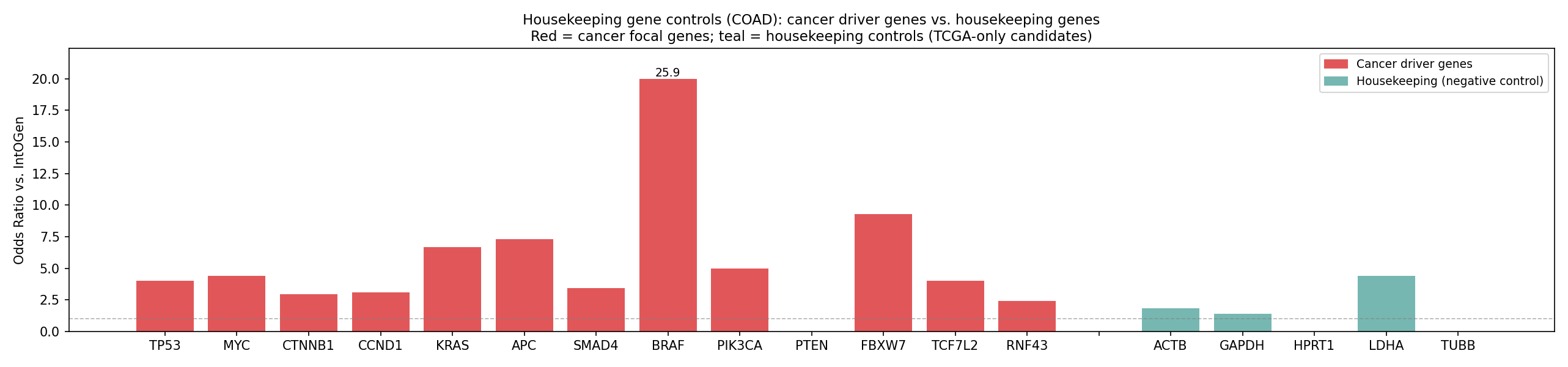}}
\caption{\textbf{(B)} COAD.}
\end{figure}

\begin{figure}
\centering
\pandocbounded{\includegraphics[keepaspectratio,alt={Figure 5. Neutral gene controls vs.~cancer focal genes. Odds ratios for TCGA-only candidates of four tumor-expressed non-driver genes (FASN, PCNA, PKM, VIM; orange) compared with the cancer focal genes (red). These genes have substantial tumor-network connectivity (8--39 TCGA-only candidates); none is nominally significant. Bars above OR = 20 are truncated and labeled with their value (BRAF in COAD, 25.9). (A) BRCA.}]{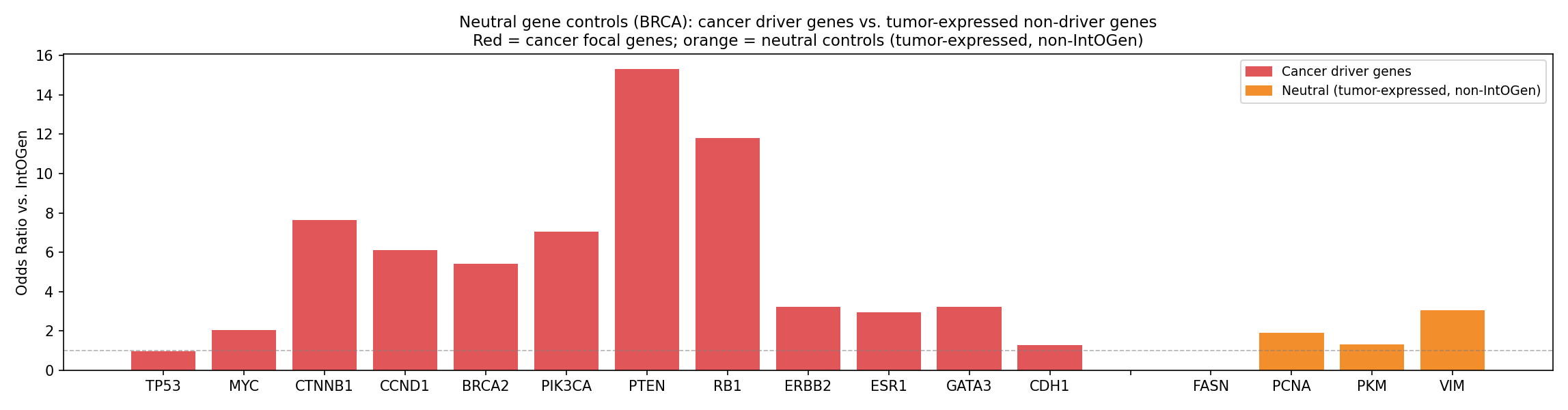}}
\caption{\textbf{Figure 5. Neutral gene controls vs.~cancer focal
genes.} Odds ratios for TCGA-only candidates of four tumor-expressed
non-driver genes (FASN, PCNA, PKM, VIM; orange) compared with the cancer
focal genes (red). These genes have substantial tumor-network
connectivity (8--39 TCGA-only candidates); none is nominally
significant. Bars above OR = 20 are truncated and labeled with their
value (BRAF in COAD, 25.9). \textbf{(A)} BRCA.}
\end{figure}

\begin{figure}
\centering
\pandocbounded{\includegraphics[keepaspectratio,alt={(B) COAD.}]{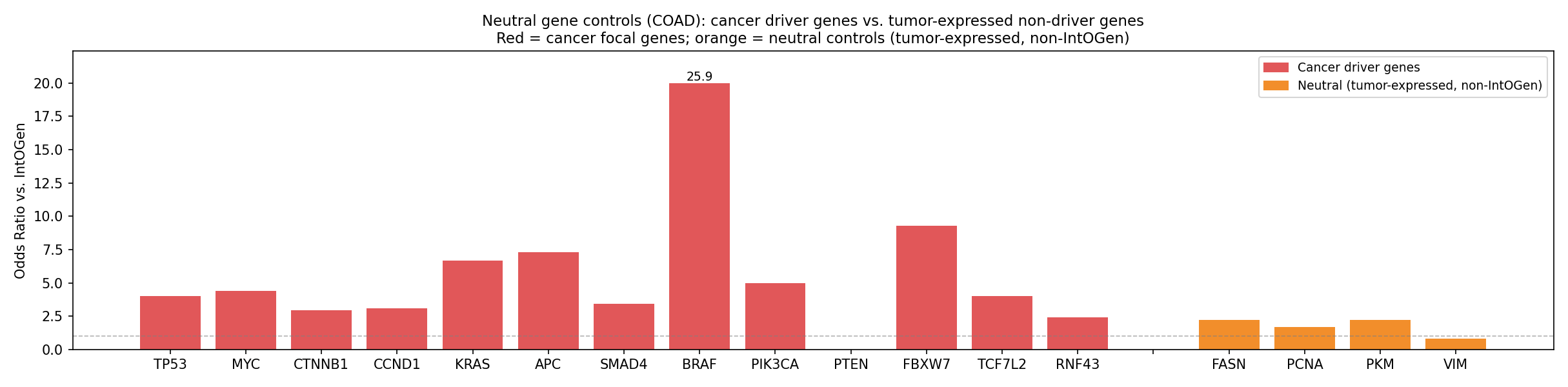}}
\caption{\textbf{(B)} COAD.}
\end{figure}

\subsection{References}\label{references}

Aibar S, et al.~(2017). SCENIC: single-cell regulatory network inference
and clustering. \emph{Nature Methods}, 14, 1083--1086.

Alvarez MJ, et al.~(2016). Functional characterization of somatic
mutations in cancer using network-based inference of protein activity.
\emph{Nature Genetics}, 48, 838--847.

Basso K, et al.~(2005). Reverse engineering of regulatory networks in
human B cells. \emph{Nature Genetics}, 37, 382--390.

Califano A, Alvarez MJ. (2017). The recurrent architecture of tumour
initiation, progression and drug sensitivity. \emph{Nature Reviews
Cancer}, 17, 116--130.

Cancer Genome Atlas Network. (2012). Comprehensive molecular
characterization of human colon and rectal cancer. \emph{Nature}, 487,
330--337.

Chakravarty D, et al.~(2017). OncoKB: A precision oncology knowledge
base. \emph{JCO Precision Oncology}, 1, 1--16.

CZI Cell Science Program, Abdulla S, Aevermann B, et al.~(2025). CZ
CELLxGENE Discover: a single-cell data platform for scalable
exploration, analysis and modeling of aggregated data. \emph{Nucleic
Acids Research}, 53(D1), D886--D900.
\url{https://doi.org/10.1093/nar/gkae1142}

Duan Y, et al.~(2016). Chromatin remodeling gene ARID2 targets cyclin D1
and cyclin E1 to suppress hepatoma cell progression. \emph{Oncotarget},
7, 45863--45875.

Giorgi FM. (2025). aracne.networks: ARACNe-inferred gene networks from
TCGA tumor datasets. R package version 1.36.0, Bioconductor 3.22.

Giorgi FM, Alvarez MJ. (2026). ARACNe-AP gene regulatory networks from
TCGA tumor datasets {[}Data set{]}. Zenodo.
\url{https://doi.org/10.5281/zenodo.22918956}

Glass K, et al.~(2013). Passing messages between biological networks to
refine predicted interactions. \emph{PLOS ONE}, 8, e64832.

Lachmann A, et al.~(2016). ARACNe-AP: gene network reverse engineering
through adaptive partitioning inference of mutual information.
\emph{Bioinformatics}, 32, 2233--2235.

LangChain AI. (2024). LangGraph Documentation.
\url{https://langchain-ai.github.io/langgraph/}

Li M, Zhao H, Zhang X, et al.~(2011). Inactivating mutations of the
chromatin remodeling gene ARID2 in hepatocellular carcinoma.
\emph{Nature Genetics}, 43, 828--829.

Li Y, Nichols MA, Shay JW, Xiong Y. (1994). Transcriptional repression
of the D-type cyclin-dependent kinase inhibitor p16 by the
retinoblastoma susceptibility gene product pRb. \emph{Cancer Research},
54, 6078--6082.

Martínez-Jiménez F, Muiños F, Sentís I, Deu-Pons J, Reyes-Salazar I,
Arnedo-Pac C, Mularoni L, Pich O, Bonet J, Kranas H, Gonzalez-Perez A,
Lopez-Bigas N. (2020). A compendium of mutational cancer driver genes.
\emph{Nature Reviews Cancer}, 20, 555--572.

Molinari F, Frattini M. (2014). Functions and regulation of the PTEN
gene in colorectal cancer. \emph{Frontiers in Oncology}, 3, 326.

Serebriiskii IG, et al.~(2022). Comprehensive characterization of PTEN
mutational profile in a series of 34,129 colorectal cancers.
\emph{Nature Communications}, 13, 1618.

Sherr CJ, Roberts JM. (1999). CDK inhibitors: positive and negative
regulators of G1-phase progression. \emph{Genes \& Development}, 13,
1501--1512.

Suehnholz SP, et al.~(2024). Quantifying the expanding landscape of
clinical actionability for patients with cancer. \emph{Cancer
Discovery}, 14, 49--65.

Xin J, et al.~(2016). High-performance web services for querying gene
and variant annotation. \emph{Genome Biology}, 17, 91.

Zhang M, Swamy V, Dupire L, Cassius R, Kanatsoulis C, Paull E,
AlQuraishi M, Karaletsos T, Califano A. (2025). GREmLN: A Cellular Graph
Structure Aware Transcriptomics Foundation Model. \emph{bioRxiv}.
\url{https://doi.org/10.1101/2025.07.03.663009}

\end{document}